\documentclass[11pt]{article}

\usepackage{acl}

\usepackage{times}
\usepackage{latexsym}

\usepackage[T1]{fontenc}

\usepackage[utf8]{inputenc}

\usepackage{microtype}

\usepackage{inconsolata}
\usepackage{graphicx}
\usepackage{tabularx}
\usepackage{colortbl}
\usepackage{tcolorbox}
\usepackage{multirow}
\usepackage{booktabs}
\usepackage{xcolor}
\usepackage[table]{xcolor}
\usepackage{amsmath}
\usepackage{adjustbox}
\usepackage{amsfonts}
\usepackage{arydshln}
\usepackage{enumitem}
\definecolor{rcolor}{gray}{0.9}

\title{Fundamental Reasoning Paradigms Induce Out-of-Domain Generalization in Language Models}

\author{ Mingzi Cao$^{\;\alpha}$\thanks{\;\;Equal contribution.}, Xingwei Tan$^{\;\alpha}$\footnotemark[1],  Mahmud Elahi Akhter$^{\;\beta}$, Marco Valentino$^{\;\alpha}$\\ \textbf{Maria Liakata}$^{\;\beta\;\gamma}$, \textbf{Xi Wang}$^{\;\alpha}$, \textbf{Nikolaos Aletras}$^{\;\alpha}$ \\
    $^\alpha$School of Computer Science, University of Sheffield\\
    $^\beta$School of Electronic Engineering and Computer Science, Queen Mary University of London\\
    $^\gamma$The Alan Turing Institute\\
    \texttt{\{mcao20,xingwei.tan,m.valentino,xi.wang,n.aletras\}@sheffield.ac.uk}\\
    \texttt{\{m.akhter,m.liakata\}@qmul.ac.uk}}

\begin{document}
\maketitle

\begin{abstract}
Deduction, induction, and abduction are fundamental reasoning paradigms, core for human logical thinking.
Although improving Large Language Model (LLM) reasoning has attracted significant research efforts, the extent to which the fundamental paradigms induce generalization has yet to be systematically explored.
In this study, we shed light on how the interplay between these core paradigms influences LLMs' reasoning behavior.
To this end, we first collect a new dataset of reasoning trajectories from symbolic tasks, each targeting one of the three fundamental paradigms, to abstract from concrete world knowledge. 
Then, we investigate effective ways for inducing these skills into LLMs. We experiment with a battery of methods including simple fine-tuning, and more complex approaches to increase model depth, or transform a dense model to a mixture-of-experts.
We comprehensively evaluate induced models on realistic out-of-domain tasks, that are entirely formulated in natural language and contain real-world knowledge. 
Our results reveal that our approach yields strong generalizability with substantial performance gains (up to $14.60$) across realistic tasks.\footnote{\;\;Our code is available at \url{https://github.com/voalmciaf/FR-OOD}.}
\end{abstract}

\section{Introduction}

Deduction, induction, and abduction are fundamental reasoning paradigms, forming the cornerstone of human logical reasoning \cite{frankfurt1958peirce,salmon1994introduction}.
\textit{Deduction} infers conclusions based on rules and observations. \textit{Induction} abstracts rules based on patterns in premises and observations, and \textit{Abduction} infers the most plausible explanation to account for an observation, given a set of rules. 
These paradigms play a crucial role in understanding and developing the reasoning capabilities of large language models (LLMs)~\citep{dougrez-lewis-etal-2025-assessing,tan-etal-2025-enhancing-logical,hu2025ahasystematicmetaabilitiesalignment}.

Previous work often conflates induction, deduction, and abduction, overlooking how they individually contribute to improving LLM reasoning~\citep{zheng-etal-2025-logidynamics,hu2025ahasystematicmetaabilitiesalignment}.
While \citet{tan-etal-2025-enhancing-logical} individually induce deductive skills via fine-tuning (FT) to improve OOD generalization, they neglect induction and abduction. Moreover, it is common practice to use data that entangles world knowledge and reasoning to improve LLM reasoning~\citep{hu2025ahasystematicmetaabilitiesalignment,guha2025openthoughtsdatarecipesreasoning,deepseekai2025deepseekr1incentivizingreasoningcapability}, making it unclear whether gains stem from genuine reasoning or factual recall. 
Consequently, two main questions remain unexplored: (1) \textit{to what extent do distinct deduction, induction, and abduction skills improve general reasoning}; and (2) \textit{what are the most effective methods for inducing these capabilities?}

\begin{figure}[!t]
\begin{center}
  \includegraphics[width=0.95\linewidth]{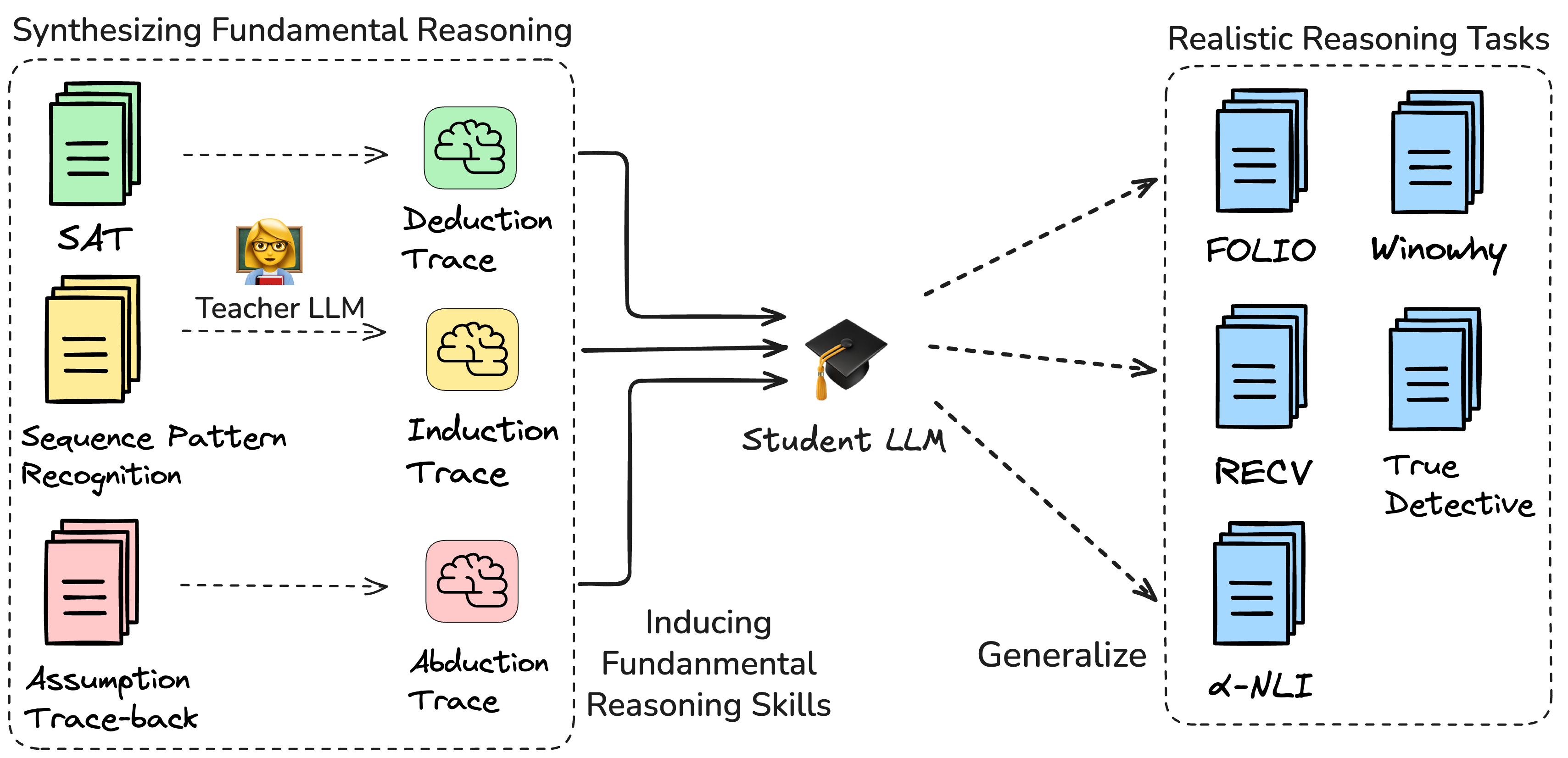} 
\end{center}
\caption{We induce fundamental reasoning skills with the trajectories generated over symbolic tasks, assessing generalization to realistic natural language tasks.}
\label{fig:introduction}
\vspace{-16pt}

\end{figure}

To answer these questions, we first enrich a procedurally-generated symbolic reasoning dataset introduced by \citet{hu2025ahasystematicmetaabilitiesalignment} with reasoning trajectories from teacher LLMs. The dataset consists of tasks, each targeting one of the three fundamental paradigms while abstracting the inference from world knowledge. However, it only consists of questions and answers without associated trajectories. The trajectories serve as demonstrations for inference patterns independent of world knowledge. Subsequently, we induce student LLMs with fundamental reasoning abilities, investigating the impact and interplay of deduction, induction, and abduction. We systematically evaluate, for the first time, a
battery of inducing methods beyond standard fine-tuning, including \emph{up-scaling} to increase model depth \citep{wu-etal-2024-llama}, allowing the effective absorption of new reasoning capabilities; and \emph{upcycling} converting standard LLMs to Mixture-of-Experts (MoE) to scale model capacity while avoiding interference between existing and newly induced information \citep{komatsuzaki2023sparse}.
Finally, we comprehensively evaluate the induced models both in-domain (i.e., on the same symbolic tasks) and Out-of-Domain (OOD) on the rest of the symbolic tasks and realistic tasks requiring real-world knowledge, across two model families (see Figure \ref{fig:introduction}). Our contributions are as follows:

\begin{itemize}

    \item We collect and release a new symbolic reasoning dataset for deduction, induction, and abduction. It contains a total of $\approx$17K symbolic problems enriched with more than 160K reasoning trajectories generated via two teacher LLMs from different families.

    \item  We show that inducing smaller student LLMs with fundamental reasoning skills, decoupled from world knowledge, yields substantial performance gains across realistic OOD tasks (up to $14.60$)  across different inducing methods and LLM families.

    \item We find that deductive abilities generally offer the highest OOD performance gains across models. We also observe strong cross-paradigm improvements, which reveal how fundamental reasoning skills in isolation  benefit LLM reasoning as a whole. On the other hand, the most effective inducing method is model-dependent, with upcycling achieving the best overall results.

\end{itemize}

\section{Related Work}

\paragraph{Eliciting Reasoning in LLMs}
Large reasoning models integrate step-by-step reasoning during post-training~\cite{olmo2025olmo3}. Recent work explores how fundamental reasoning paradigms impact the effectiveness of LLM reasoning.
\citet{zheng-etal-2025-logidynamics} evaluate different logical inference paradigms, but they only comparing inductive reasoning against the combination of abduction and deduction, lacking understanding of individual fundamental reasoning paradigms. 
\citet{hu2025ahasystematicmetaabilitiesalignment} separately specialize LLMs on deductive, inductive, and abductive reasoning, and subsequently merge the resulting models, conflating all reasoning paradigms together. This makes it difficult to isolate how each individual fundamental reasoning paradigm impacts performance.
Therefore, we learn each specific fundamental reasoning paradigm individually via training on reasoning trajectories collected by teacher LLMs. This approach provides a clearer and fine-grained understanding of their impact on LLM reasoning for the first time.

\paragraph{Synthetic Reasoning Trajectories}
LLM reasoning requires large-scale trajectories in post-training \citep{liu2024datasetslargelanguagemodels,yang2025qwen3technicalreport}. However, human-curated reasoning data is often noisy \citep{NEURIPS2024_8678da90}, and difficult to scale \citep{sun-etal-2024-rapidly}. Therefore, synthetic reasoning trajectories from teacher models are increasingly used to train student models \citep{deepseekai2025deepseekr1incentivizingreasoningcapability,olmo2025olmo3}. While they offer substantial gains in general \citep{guha2025openthoughtsdatarecipesreasoning} and specialized domains \citep{yu2024metamath,zhao2025agentardeepfinance100klargescalefinancialdataset,yu2025finemedlmo1enhancingmedicalknowledge}, the  trajectories include world knowledge, making it difficult to isolate fundamental reasoning improvements. \citet{tan-etal-2025-enhancing-logical} use symbolic reasoning trajectories to decouple world knowledge but focus exclusively on deductive reasoning, leaving other fundamental reasoning paradigms unexplored.

\begin{figure}[t]
    \centering
    \includegraphics[width=0.95\linewidth]{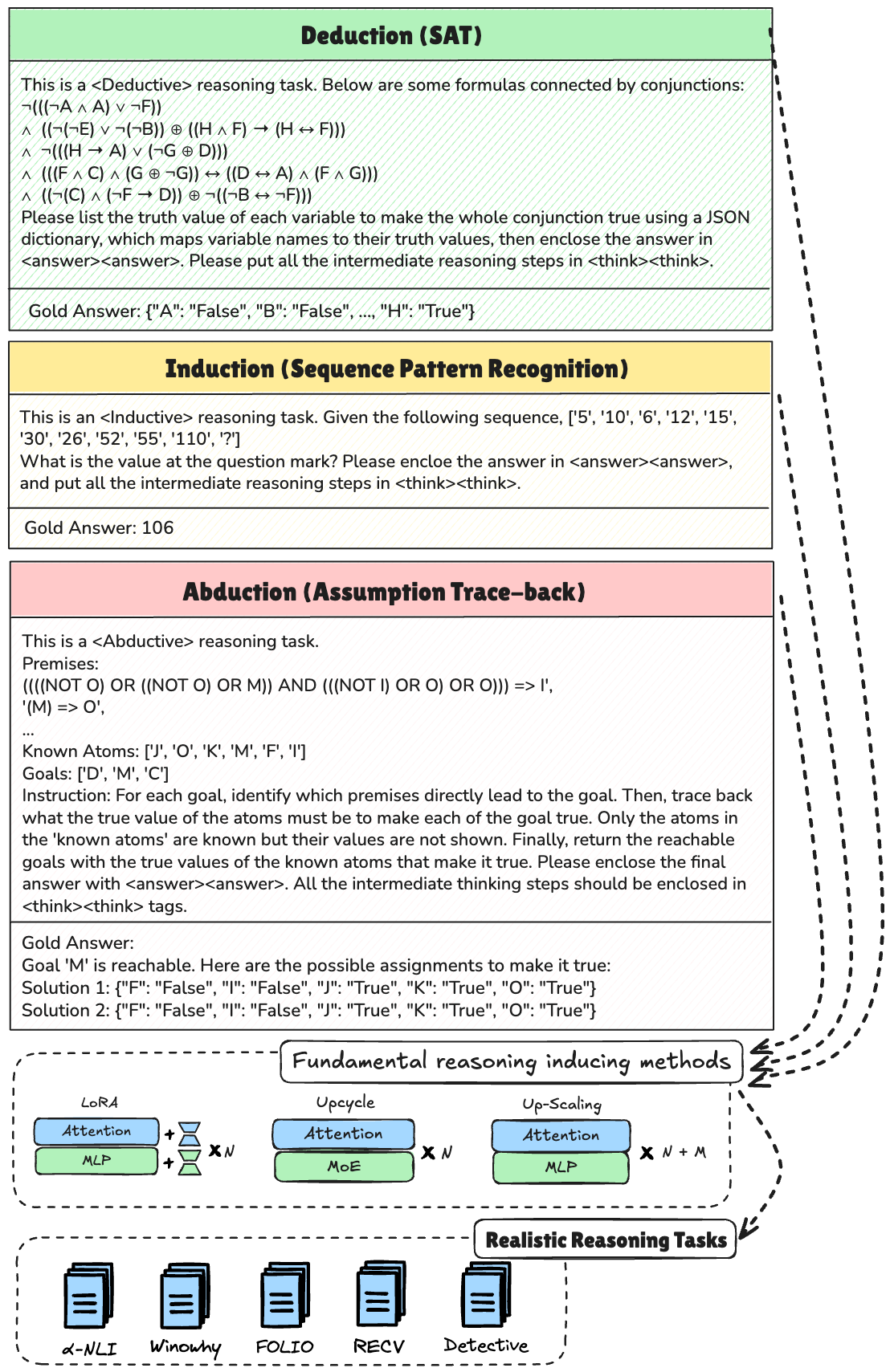}
    \caption{Symbolic reasoning tasks, and the fundamental reasoning inducing process.}
    \label{fig:reasoning_tasks}
\end{figure}

\section{Fundamental Reasoning Data}

Our aim is to investigate the impact of inducing LLMs with the three fundamental reasoning types: deduction, induction, and abduction. For this purpose, we need corresponding trajectories.

\subsection{Symbolic Reasoning Tasks}
\label{sec:our_tasks}

We enrich an existing fundamental reasoning dataset developed by \citet{hu2025ahasystematicmetaabilitiesalignment}. We opted for this data because it is publicly available and comprises 3,600 deductive, 4,500 inductive, and 9,000 abductive questions with their answers, but without associated reasoning trajectories that are essential to our study. The data is procedurally-generated based on three symbolic tasks (Figure \ref{fig:reasoning_tasks}). 

\paragraph{Deduction.} The task consists of Boolean Satisfiability (SAT) questions, requiring the truth value of the variables to satisfy a given formula. All problems are guaranteed to be satisfiable. The original questions list the formulas separately without using a Conjunctive Normal Form (CNF), i.e., a standardized way to write Boolean formulas as a conjunction of clauses, where each clause is a disjunction of variables or their negations. Our preliminary experimentation showed that this creates confusion for LLMs. Furthermore, the original answers are presented in plain text, which complicates automated evaluation. Therefore, we modify the original formulas into CNFs and express the answers in JSON for collecting reasoning trajectories.

\paragraph{Induction.} This task requires the model to predict the next number in a given sequence of numbers, representing a basic form of inductive reasoning. The numerical sequences consist of repeating patterns that involve addition, subtraction, or multiplication. We use the original input-output pairs without any modifications.

\paragraph{Abduction.} The task uses logic formulas and known atoms with hidden truth values. The objective is to first identify all reachable goals from the atoms by checking if formulas related to the goals exist. 
The model requires abductive reasoning to work backward from an assumed conclusion (i.e., the goals are true) to hypothesizing the most likely set of premises (antecedent truth values). However, the original task instructions do not clarify that known atoms do not necessarily equate to true atoms, leading to misinterpretations by LLMs. Moreover, the original answers only check for goal reachability despite requiring LLMs to identify related premises and trace back the truth values of the known atoms. This oversimplification causes the model to simply check whether the goals appear in the premises. To overcome these limitations, we keep the original formulas but modify the task instructions and the answers by searching the possible truth values of known atoms that make the reachable goal true using Prolog \cite{wielemaker:2011:tplp}. Table ~\ref{tab:abductive_example} (Appendix~\ref{app:data_examples}) shows a comparison between the original and our abductive task.

\subsection{Reasoning Trajectories Sampling}
\label{sec:our_data}

\paragraph{Data Splits.}
From each task type, we use 100 questions and answer pairs for testing and 100 for development, reserving the remaining for training. We collect reasoning trajectories only for the training set.
A question-answer pair and its trajectories only appear in one of the splits.

\paragraph{Teacher Models.} We sample reasoning trajectories using two widely used open-weight LLMs: \textit{Qwen3-30B-Instruct} \citep{yang2025qwen3technicalreport} and \textit{Llama-3.3-70B-Instruct} \citep{grattafiori2024llama3herdmodels}.\footnote{We use the vLLM API \citep{kwon2023efficient}, ensuring efficient inference.}

\paragraph{Sampling.} 
Questions are fed to a teacher LLM  (see Appendix \ref{app:data_examples} for prompts) in a zero-shot setting.
For each question, we sample five responses using different random seeds. We set the maximum number of generated tokens to 10,000, ensuring the trajectories reach the conclusion.  We exclude responses with fewer than 20 words since such short texts are unlikely to contain valid reasoning trajectories. For the Qwen model, this results in filtering out 554 deductive, 2,430 inductive, and 19,872 abductive trajectories. However, no trajectories from the Llama model were filtered out.
We find that the filtered responses of Qwen are either empty or strings including random tokens, perhaps generated due to confusion.
Due to the fact that automatically verifying the logic consistency of textual reasoning trajectories remains an open problem and LLMs learn more from structure instead of content \cite{li-etal-2025-language-models}, we do not filter out more trajectories.

The final training set comprises $\sim$16,000  questions together with $\sim$82,000 and $\sim$60,000 trajectories generated by Llama and Qwen teacher models, respectively. Table \ref{tab:dataset_stats} shows detailed statistics. We note that Qwen generates longer trajectories than Llama, particularly on inductive and abductive tasks (four times longer on average).
We further observe that Qwen tends to repeat formulas and sequences from the questions in its reasoning steps. It also frequently uses tables to summarize its responses and generates many special tokens such as ``<tool\_call>'', ``<IActionResult]'', and ``<pal>''.
This is likely due to post-training on tool calling.

\begin{table}[!t]
\centering
\tiny
\resizebox{0.95\columnwidth}{!}{%
\begin{tabular}{clrrrrr}
\toprule
\textbf{Teacher} &\textbf{Type} & \textbf{\# Quest.} & \textbf{\# Traject.} & \textbf{\# Tokens} & \textbf{Avg. Tokens} \\
\midrule
& De. & 3,400 & 17,000 & 18.6M & 1,093\\
\multirow{-3}{*}{Llama3.3-70B-IT} & In. & 4,300 & 21,500 & 11.7M & 544\\
& Ab. & 8,800 & 44,000 & 51.2M & 1,164\\
\midrule
& De. & 3,400 & 16,446 & 23.0M & 1,397\\
\multirow{-3}{*}{Qwen3-30B-IT} & In. & 4,300 & 19,070 & 47.8M & 2,507\\
& Ab. & 8,800 & 24,128 & 135.9M & 5,631\\
\bottomrule
\end{tabular}%
}
\caption{Fundamental reasoning training data statistics.}
\vspace{-16pt}

\label{tab:dataset_stats}
\end{table}

\section{Inducing Fundamental Reasoning Skills}

We experiment with various inducing methods, from standard FT to more sophisticated approaches that introduce structural changes to the model.

\paragraph{Full FT.}

First, we simply update all model parameters via next-token prediction on the collected reasoning trajectories, similar to \citet{tan-etal-2025-enhancing-logical}.

\paragraph{LoRA FT.}
As an efficient alternative, we also experiment with Low-Rank Adaptation \citep[LoRA]{hu2022lora} FT.
Instead of updating an existing weight matrix $W \in \mathbb{R}^{d \times k}$, LoRA injects a low-rank decomposition $\Delta W = B A$, where $A \in \mathbb{R}^{r \times k}$ and $B \in \mathbb{R}^{d \times r}$, updating only $A$ and $B$.

\paragraph{Up-scaling.}
LoRA's low-rank constraint while efficient, limits its effectiveness in domain adaptation \citep{biderman2024lora, pletenev-etal-2025-much}. Model up-scaling \citep{wu-etal-2024-llama, cao2025progressivedepthupscalingoptimal} increases model capacity by inserting additional layers into a pre-trained LLM. By updating only the new layers, up-scaling preserves original parametric knowledge. It also allows better acquisition of new knowledge for language  \citep{choudhury2025llama3nanda10bchatopengenerativelarge,lee-etal-2025-enginius} and domain adaptation~\citep{hoffmann2025llamagenba10btrilinguallargelanguage, lim2025expandingfoundationallanguagecapabilities}. We up-scale models with LLaMAPro \citep{wu-etal-2024-llama} by dividing base models into multiple layer groups and insert new layers atop each group. The new layers are initialized by copying weights from the corresponding base layers, with the attention and MLP output projection matrices set to zero.

\paragraph{Upcycling.}
Upcycling \citep{komatsuzaki2023sparse, nakamura2025dropupcycling} transforms a dense Transformer layer to a MoE layer by expanding the MLP block into multiple expert MLPs. During inference, a router chooses experts. It has been successfully used for multilingual \citep{10.1609/aaai.v39i24.34805, zhang-etal-2025-less} and domain adaptation \citep{li2025dynamicexpertspecializationcatastrophic}. 
Compared to the depth expansion of up-scaling, upcycling is an efficient width expansion approach. We use Sparse-Upcycling \citep{komatsuzaki2023sparse} as it does not require additional training for the router \citep{gritsch-etal-2025-nexus}, or extra modules beyond the MoE structure \citep{10.1609/aaai.v39i24.34805}. Sparse-Upcycling initializes new expert MLPs by copying the original pre-trained MLP weights, with the router randomly initialized.

\section{Experimental Setup}

\subsection{Training Settings}
To study how each of the fundamental reasoning paradigms influence reasoning behavior, we train LLMs separately using \textit{Deduction}, \textit{Induction}, and \textit{Abduction} reasoning trajectories respectively. For comparison, we also induce LLMs with all three reasoning types together (i.e., \textit{Mix}) for assessing whether they offer complementary inductive bias. 

\subsection{Models \& Baselines}
We use two student models: \textit{Llama-3.1-8B-Instruct} \citep{grattafiori2024llama3herdmodels} with 32 layers; and \textit{Qwen3-8B} \citep{yang2025qwen3technicalreport} with 36 layers. 

We compare inducing methods to (1) student models before FT (\textit{vanilla}); and (2) \textit{teacher models} used for collecting the trajectories (\S\ref{sec:our_data}).

\subsection{Implementation Details} 
For fair comparison, we keep the parameters of all induced models at a similar size, updating only the newly-added parameters, except of Full FT. We apply LoRA to all linear layers of the students, with ranks set to 1,024 for \textit{Llama-3.1-8B-Instruct} and 512 for \textit{Qwen3-8B}. For up-scaling, we divide each student into four layer groups, inserting three new layers into each group. We upcycle odd-numbered dense layers into MoE layers by adding a single expert MLP per layer, following \citet{JMLR:v23:21-0998}. 
Further details are in Appendix \ref{sec:imp_detail}.

\begin{figure*}[!t]
    \centering
    \includegraphics[width=0.95\linewidth]{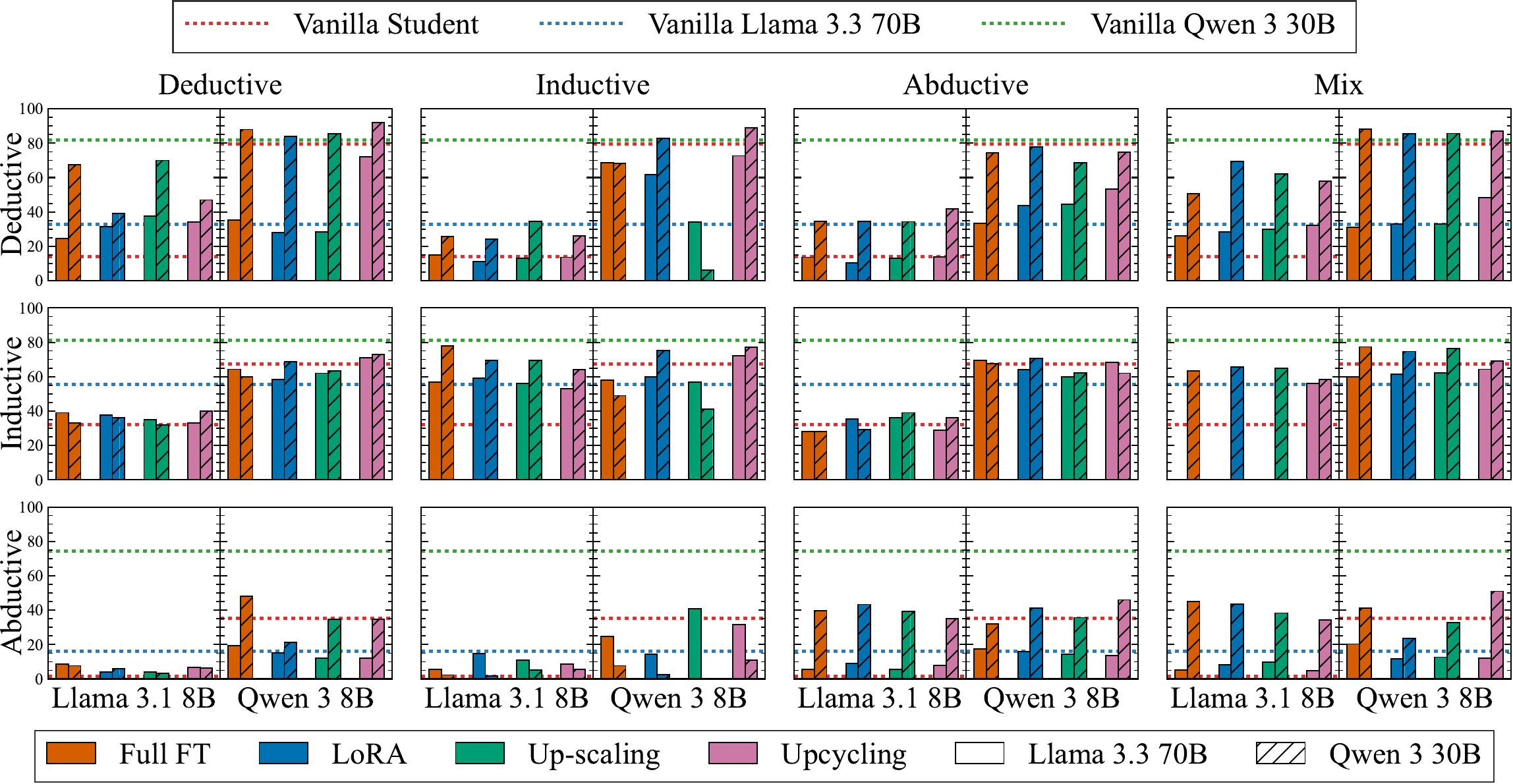}
    \caption{Symbolic task accuracy of induced models across different settings. X-axis and y-axis denote inducing paradigms and symbolic test tasks, respectively. Solid/striped bars denote teacher for generating trajectories. Dotted lines denote performance for the vanilla student and teacher models.}
    \label{fig:symbolic_performance}
\end{figure*}

\subsection{Evaluation}

\paragraph{Symbolic In-Domain.}
We evaluate on the test set (\S \ref{sec:our_data}) that corresponds to the reasoning type used for training (e.g., training and testing on inductive tasks). When training on the Mix dataset, the test sets of all three reasoning types are in-domain.\looseness=-3

\paragraph{Symbolic OOD.} To test symbolic generalization, we evaluate on symbolic tasks not seen during training (e.g., train on abductive,  test on deductive).\looseness=-3

\paragraph{Realistic OOD.}
We include the following realistic OOD datasets: \textit{True Detective} \citep{del-fishel-2023-true} contains detective puzzles with multi-choice questions; $\alpha NLI$ \citep{zhao-etal-2023-abductive} consists of multi-choice questions of the most probable explanation given a context; \textit{WinoWhy} \citep{zhang-etal-2020-winowhy} includes explanations for the questions in Winograd Challenge \citep{10.5555/3031843.3031909}, aiming to distinguish plausible reasons from  similar but wrong reasons; \textit{FOLIO} \citep{han-etal-2024-folio} contains logical reasoning questions in natural language. Given the context, a set of first-order logic formulas, the model predicts if a statement is true, false, or uncertain; \textit{RECV} \citep{dougrez-lewis-etal-2025-assessing} contains claims from Twitter and Wikipedia, for determining whether they are true or false, requiring deductive or abductive reasoning.\looseness=-3 

\paragraph{Metrics.} We report the accuracy of the final answer, judged by \textit{Qwen3-30B-Instruct}. 
We use this judge based on recent analysis by \cite{haldar-hockenmaier-2025-rating}, which indicates it is more robust than competitive alternatives (including Llama-3 and Deepseek-R1~\citep{deepseekai2025deepseekr1incentivizingreasoningcapability}) in terms of coherence, consistency, and relevance.\looseness=-3

\begin{figure*}[!t]
    \centering
    \includegraphics[width=0.95\linewidth]{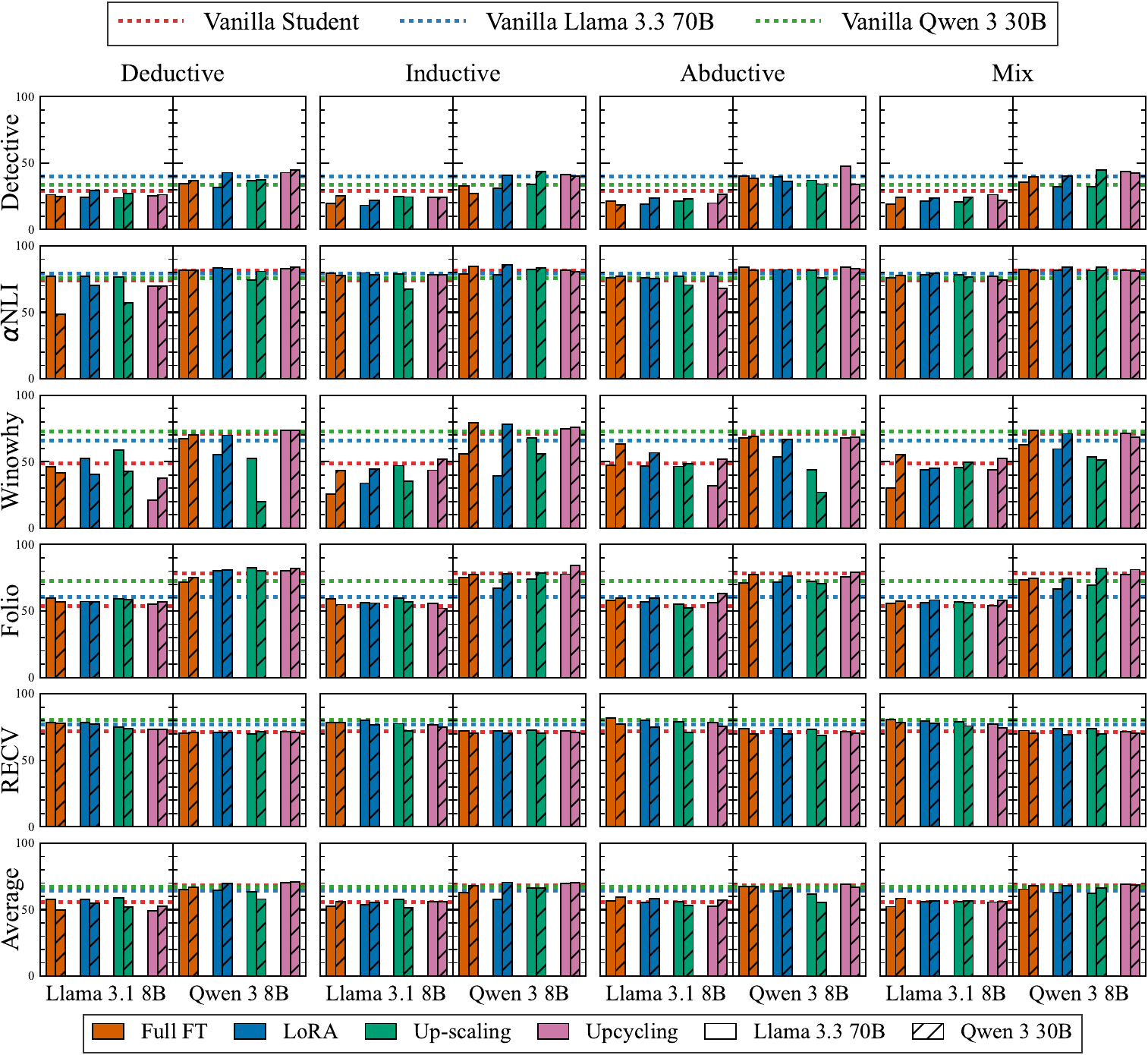}
    \caption{Realistic OOD task performance when inducing student model across different settings.}
    \label{fig:OOD_G}
\end{figure*}

\section{Results}
Figures~\ref{fig:symbolic_performance} and~\ref{fig:OOD_G} show model  performance on symbolic and realistic tasks respectively. Tables~\ref{tab:results} and~\ref{tab:qwen_results} (Appendix~\ref{sec:Full_Results}) show full results and std deviations. 

\subsection{Symbolic Tasks}
\label{sec:In-Domain}

\paragraph{In-Domain: Specialized training yields the highest gains.}
The subplots on the main-diagonal and the last column in Figure~\ref{fig:symbolic_performance} represent in-domain settings, where the train and test paradigms match. In five of the six scenarios (2 student models $\times$ 3 symbolic tasks), specialist induced models substantially outperform the vanilla student. 
We find that inducing \textit{Deduction} yields the largest improvements (e.g., \textit{Llama-3.1-8B} gains $56.00$ via up-scaling, while \textit{Qwen-3-8B} is up $12.33$ via upcycling). 
\textit{Induction} follows ($46.00$ via Full FT, $9.67$ via Upcycling), while inducing \textit{Abduction} produces the smallest comparative gains ($41.66$ via LoRA, $10.67$ via Upcycling). 
One possible explanation of the larger deduction gains  is that the transformer architecture has the right structure for deduction \cite{aggarwal2025bayesiangeometrytransformerattention}.
Notably, the \textit{Mix} setting, which trains LLMs on a mixture of all three paradigms, offers a competitive inducing strategy. While the \textit{Mix} setting does not yield the largest in-domain performance gains achieved by single-paradigm specialists, it still outperforms the vanilla \textit{Qwen-3-8B} on all symbolic tasks in six out of eight cases (2 teachers $\times$ 4 inducing methods).

\paragraph{OOD: Asymmetric transfer between paradigms.}
\label{sec:symbolic_OOD}
Turning to the other subplots in Figure~\ref{fig:symbolic_performance}, we analyze OOD generalization, where models trained on one paradigm are tested on other. We observe a distinct order of effectiveness: \textit{Deduction > Induction > Abduction}. Gains in deduction transfer strongly to other tasks. Across 16 symbolic cases per paradigm, deduction gains transfer in 12 cases, improving induction and abduction in nine cases each. In contrast, induction transfers moderately (co-occurring with gains in 7 cases), while abduction transfers poorly, benefiting deduction in only four cases and induction in five cases. For instance, when inducing \textit{Qwen-3-8B} via upcycling, deduction yields the most transferable gains ( induction: $73.00$, abduction: $34.67$). Induction achieves moderate transfer, exhibiting asymmetric generalization (deduction: $89.00$, abduction: $11.00$). In contrast, inducing abduction does not provide comparable gains (deduction: $74.67$, induction: $62.00$). 
These results show the transfer between fundamental reasoning paradigms is asymmetric. While abduction is the hardest to learn (lowest in-domain gains), deduction is the most transferable skill, serving as a \textit{universal donor} for reasoning capabilities.

\paragraph{Inducing methods: Capacity expansion vs.\ efficient adaptation.}
Looking at different inducing methods, we observe that the most effective method is model-dependent, though upcycling generally offers more robust gains. 
For \textit{Qwen3-8B}, upcycling consistently offers the highest gains across paradigms (e.g., $12.33$ in deduction, $9.67$ in induction and $10.67$ in abduction). In contrast, \textit{Llama-3.1-8B} benefits from a diverse range of inducing methods. Up-scaling drives its peak gains in deduction ($56.00$) and Full FT dominates in induction ($46.00$). While LoRA is a competitive efficient alternative, surprisingly achieving the best abduction improvement ($41.66$), it generally lags capacity-expanding methods (upcycling), suggesting that fundamental reasoning skills benefit from the additional capacity offered by structural adaptation.

\paragraph{Stronger teachers lead to higher student performance.}
\textit{Qwen-3-30B} substantially outperforms \textit{Llama-3.3-70B} on symbolic tasks (deduction: $82.00$ vs.\ $33.00$, induction: $81.00$ vs. $55.33$, abduction: $74.33$ vs. $16.00$). Both student LLMs benefit more from the stronger Qwen teacher for symbolic tasks. 
For example, \textit{Qwen3-8B}, using the Llama teacher, causes performance drops in most cases (43/48). When shifting to Qwen teacher, its deduction performance substantially improves (up to $92.00$ via deduction upcycling) even surpassing the teacher model ($82.00$). Such improvements on the deduction task are observed across different inducing paradigms (except abduction), further supporting the asymmetric transfer  between paradigms.
The performance difference between teachers highlights that the quality of reasoning traces is crucial for symbolic generalization.

\subsection{Realistic Tasks}
\label{sec:OOD}

\paragraph{Deduction > Induction > Abduction.}
\label{sec:OOD_order}

We observe that the order of effectiveness established in the symbolic experiments (\S\ref{sec:symbolic_OOD}) persists in realistic tasks. As shown in Figure~\ref{fig:OOD_G}, inducing deduction offers the strongest realistic OOD performance. For example, both \textit{Qwen-3-8B} and \textit{Llama-3.1-8B} achieve the highest average performance when inducing deduction ($70.14$ via upcycling, $58.70$ via up-scaling) using traces from the Llama teacher. Induction is slightly lower ($69.49$ via upcycling, $57.58$ via up-scaling), while abduction yields more modest gains ($69.34$ via upcycling, $56.79$ via Full FT). The \textit{Mix} strategy is worse than any single-paradigm approach ($69.04$ via upcycling, $56.07$ via up-scaling), contrary to the symbolic performance. We hypothesize that, unlike symbolic tasks, realistic tasks are less structured, causing the \textit{Mix} setting to dilute the strong inductive bias provided by single-paradigms, especially deduction. 

\paragraph{Knowledge-free inducing goes beyond the teacher.}
Inducing reasoning paradigms, abstracted from world knowledge, enables student LLMs to surpass their teachers even in realistic tasks. For example, when training Llama student (RECV: $71.98$) with Llama teacher (RECV: $77.07$), 13 out of 16 induced models (four inducing paradigms $\times$ four inducing methods) outperform the teacher, achieving up to $81.73$ via abductive Full FT. Strikingly, this improvement remains observable even when the student initially outperforms the teacher. For example, when training Qwen student ($\alpha$NLI: $81.26$) with Qwen teacher ($\alpha$NLI: $75.77$), the student performs better in the majority of cases (13/16), reaching up to $84.24$ via inductive upcycling. Such improvements are observed across tasks under different teacher–student pairs. We attribute these gains to the knowledge-free symbolic trajectories. The teacher transfers better fundamental reasoning skills to the student through symbolic structures, avoiding the content bias of world knowledge, pervasive in real tasks.

\begin{table*}[t]
\tiny
\centering
\resizebox{0.81\textwidth}{!}{%
\begin{tabular}{l p{0.26\textwidth} p{0.26\textwidth} p{0.26\textwidth}}
\toprule

\textbf{Case 1}
& \multicolumn{3}{p{0.78\textwidth}}{\textbf{Premises:}
(1) People in this club who perform in school talent shows often attend and are very engaged with school events.
$\ldots$
(3) People in this club who chaperone high school dances are not students who attend the school.
$\ldots$
(6) Bonnie is in this club and she either (i) both attends and is very engaged with school events and is a student who attends the school, or (ii) is not someone who both attends and is very engaged with school events and is not a student who attends the school.
} \\
\addlinespace

\textbf{Conclusion} 
& \multicolumn{3}{p{0.78\textwidth}}{Bonnie performs in school talent shows often.} \\
\midrule

\textbf{Model} 
& \textbf{Vanilla Llama} 
& \textbf{Deductive Full FT Llama} 
& \textbf{Deductive Up-scaling Llama} \\
\midrule

\textbf{Reasoning} 
& \textit{``1. If Bonnie is a student... then she must perform in school talent shows often...
5. This means that if Bonnie is not a student... then she must chaperone high school dances, \textbf{which contradicts the fact that she is not a student}''} 
& \textit{``...\textbf{we cannot definitively say if she performs in school talent shows often}......we have a premise that supports this conclusion \textbf{if she is a student}, we can say that the conclusion is True \textbf{if we assume Bonnie is a student}...''} 
& \textit{``...cannot directly conclude that Bonnie performs in school talent shows often... premise 6 does not directly state that Bonnie is a student or that she is not a student. It only gives two possible scenarios for Bonnie's situation... \textbf{cannot be certain of it}.''} \\
\addlinespace

\textbf{Analysis} 
& \textbf{\textcolor{red}{Directionality Error:}} Illegally reverses implication ($P \rightarrow Q$ as $Q \rightarrow P$) and hallucinates a contradiction to force a determinate conclusion. 
& \textbf{\textcolor{red}{Conditional Reasoning Drift:}} Recognizes uncertainty but nonetheless resolves it via assumption-dependent reasoning. 
& \textbf{\textcolor{blue}{Disjunction Preservation:}} Correctly treats the disjunction in Premise 6 as epistemic uncertainty and maintains logical validity. \\
\midrule

\textbf{Final Answer} 
& \texttt{True} 
& \texttt{True} 
& \texttt{Uncertain} \\

\midrule\midrule
\textbf{Case 2} 
& \multicolumn{3}{p{0.78\textwidth}}{\textbf{Premises:} 
(1) Barutin Cove is a cove named after the Bulgarian settlement of Barutin.
(2) Barutin Cove is on the southwest coast of Snow Island.
(3) Snow Island, Greenwich Island, and Deception Island are located in the South Shetland Islands.
(4) Antarctica is located on the South Shetland Islands.
(5) If place A is located in place B and place B is located in place C, then place A is located in place C.
} \\ 
\addlinespace

\textbf{Conclusion} 
& \multicolumn{3}{p{0.95\textwidth}}{Barutin Cove is not located in Antarctica.} \\
\midrule

\textbf{Model} 
& \textbf{Vanilla Qwen} 
& \textbf{Abductive Upcycling Qwen} 
& \textbf{Deductive Upcycling Qwen} \\
\midrule

\textbf{Reasoning} 
& \textit{``...the South Shetland Islands are not part of Antarctica, but rather located near Antarctica... which seems to be a contradiction...''} 
& \textit{``...there's a potential confusion in the premises... Antarctica is a continent, while the South Shetland Islands are a group of islands that are geographically close to Antarctica...''} 
& \textit{``...So maybe the premise is incorrect? But I have to take the premises as given, even if they might be factually wrong...''} \\
\addlinespace

\textbf{Analysis} 
& \textbf{\textcolor{red}{World-Knowledge Bias:}} Overrides given premises using real-world knowledge, invents a spurious contradiction, and derives a formally correct answer by invalidating the logical setting. 
& \textbf{\textcolor{red}{Plausibility Reinterpretation:}} Treats premises as revisable observations, resolves tension via plausibility-based reinterpretation, and violates premise-faithful logical entailment. 
& \textbf{\textcolor{blue}{Premise-Faithful Deduction:}} Explicitly distinguishes logical validity from factual correctness, enforces strict premise adherence, and rejects the conclusion through rule-consistent inference. \\
\midrule

\textbf{Final Answer} 
& \texttt{True} 
& \texttt{True} 
& \texttt{False} \\

\bottomrule
\end{tabular}
}
\caption{Qualitative comparison of reasoning traces on realistic tasks across models before and after inducing fundamental reasoning skills. Full example for Case 1 and Case 2 are included in Appendix \ref{app:ood-example}.
}
\label{tab:qualitative_logic_complete}
\end{table*}

\paragraph{Cross-paradigm transfer: The Abduction Paradox.}
Inducing the reasoning paradigm corresponding to a realistic task does not always yield best performance. This is evident in realistic tasks that rely heavily on abduction. For \textit{True Detective}, training on deduction yields the best performance for both \textit{Llama-3.1-8B} and \textit{Qwen3-8B} ($26.18$ and $44.50$). In contrast, training on abduction performs the worst ($21.47$ and $38.57$). A similar pattern emerges on $\alpha$NLI, where inducing induction ($79.73/85.76$) consistently outperforms abduction ($77.37/82.73$). Since abduction is inherently difficult to learn, models benefit more from the transferable logic acquired via deduction or induction.\looseness=-1

\paragraph{Task-specific concentration of gains.}
OOD generalization is highly concentrated on specific tasks. Compared to the vanilla model, gains are higher on $\alpha$NLI (improving in 50 out of 64 cases) and Folio (41/64 cases). However, the source of these gains varies by teacher family. Llama teachers are more effective on RECV (28/32 cases), whereas Qwen teachers struggle (16/32) but perform better on WinoWhy (increasing cases from 5 to 13). This shows that while the training paradigm decides OOD effectiveness, teacher models determine which tasks benefit from the transfer.\looseness=-1

\paragraph{Students learn better from teachers from the same family.}
While symbolic tasks benefit from the \textit{stronger} Qwen teacher regardless of model family, realistic OOD generalization improves when the teacher and student belong to the \textbf{same model family}, contradicting our findings in symbolic tasks (\S\ref{sec:In-Domain}). For example, when inducing deduction for \textit{Llama-3.1-8B}, the best performance on realistic tasks is achieved with the Llama teacher ($58.70$) rather than the Qwen teacher ($54.79$). This pattern is consistent across both student models when inducing deduction or induction. We speculate that realistic tasks are more sensitive to stylistic mismatches inherent in cross-family distillation. This aligns with prior work on cross-model trajectory mismatch \citep{wen2025reasoningscaffoldingdistillingflow},  suggesting that, minimizing distributional shift between teacher and student is more critical than teacher capability.

\paragraph{Inducing Methods: Consistent with symbolic tasks.}
We observe a similar pattern to the symbolic domain (\S\ref{sec:In-Domain}): although the most effective inducing method is model-dependent, upcycling provides more robust improvements overall. Across all eight settings (two teacher models $\times$ four inducing paradigms), upcycle \textit{Qwen-3-8B} achieves the highest realistic performance in seven out of eight cases. For example, with Llama teacher, upcycling consistently yields the best realistic performance for \textit{Qwen-3-8B} across all paradigms (deduction: $70.14$, induction: $69.49$, abduction: $69.34$, \textit{Mix}: $69.04$). In contrast, \textit{Llama-3.1-8B} benefits from a diverse range of inducing methods (i.e. Full FT, LoRA, Up-scaling). This observation shows the consistency of OOD generalization across symbolic and realistic tasks when the same inducing method is applied. Furthermore, the superior performance of upcycling further illustrates that more sophisticated inducing methods outperform the current practice on using simpler Full FT, i.e., \citet{tan-etal-2025-enhancing-logical}.

\section{Qualitative Analysis}

Table~\ref{tab:qualitative_logic_complete} shows examples of reasoning behavior using different inducing methods. In Case 1, the vanilla Llama and deductive Full FT Llama fail by prematurely resolving epistemic uncertainty, committing to a determinate conclusion through implicit assumption-making. In contrast, the deductive up-scaling Llama succeeds by preserving the uncertainty deriving from the disjunctive premise. This indicates that stronger structural constraints introduced by inducing methods support logical validity, preventing overconfident conclusions on underdetermined realistic reasoning problems.

In Case 2, the vanilla Qwen and abductive upcycling Qwen fail by prioritizing world knowledge over logical reasoning, revising implausible premises and overriding the given logical rules. Conversely, the deductive upcycling Qwen succeeds by strictly adhering to the stated premises and applying logical rules regardless of content. This behavior aligns with the intrinsic nature of abduction and deduction. Abduction focuses on exploring the most plausible observations under a set of rules and thus encourages premise reinterpretation. In contrast, deduction derives conclusions from rules and observations and inherently promotes premise faithfulness. This further emphasizes how inducing reasoning paradigms decoupled from world knowledge reduce content biases, leading to more robust reasoning generalization on realistic tasks. We further include an analysis on symbolic tasks, detailed in Appendix~\ref{app:id_example}.

\section{Conclusion \& Takeaways}

We systematically studied how inducing fundamental reasoning types (i.e., deduction, induction, and abduction) shapes the generalization of LLM reasoning. By training on reasoning trajectories abstracted from world knowledge, we isolate the effect of each paradigm. We evaluate a range of inducing methods, from standard fine-tuning to capacity expansion approaches such as up-scaling and upcycling. Our results show that knowledge-free inducing yields substantial and transferable OOD gains. \textit{Deduction is the most robust paradigm and upcycling provides the largest overall improvements, enabling student LLMs to acquire structural inductive biases that even surpass their teachers on realistic tasks.}
Future work will extend this framework and explore principled ways of combining multiple reasoning paradigms without diluting their individual contribution.


\section*{Limitations}

All data and evaluations are in English given the availability and diversity of reasoning datasets. We acknowledge that the linguistic expression and underlying cognitive strategies for fundamental reasoning may differ across languages which is an interesting avenue for future work.

In addition, while many different symbolic tasks can be generated to represent the three fundamental reasoning paradigms, this work explored a single representative task for deduction, induction, and abduction, respectively. We focused, in fact, on the interplay between fundamental paradigms and inducing methods, prioritising depth of analysis rather than exhaustive coverage of the whole spectrum of symbolic reasoning tasks. While we found the proposed tasks to be sufficient to induce strong OOD generalisation, future work might investigate the potential impact of different symbolic data on performance and generalisation.


\bibliography{anthology,custom}

@misc{hu2025ahasystematicmetaabilitiesalignment,
      title={Beyond 'Aha!': Toward Systematic Meta-Abilities Alignment in Large Reasoning Models}, 
      author={Zhiyuan Hu and Yibo Wang and Hanze Dong and Yuhui Xu and Amrita Saha and Caiming Xiong and Bryan Hooi and Junnan Li},
      year={2025},
      eprint={2505.10554},
      archivePrefix={arXiv},
      primaryClass={cs.CL},
      url={https://arxiv.org/abs/2505.10554}, 
}

@inproceedings{han-etal-2024-folio,
    title = "{FOLIO}: Natural Language Reasoning with First-Order Logic",
    author = "Han, Simeng  and
      Schoelkopf, Hailey  and
      Zhao, Yilun  and
      Qi, Zhenting  and
      Riddell, Martin  and
      Zhou, Wenfei  and
      Coady, James  and
      Peng, David  and
      Qiao, Yujie  and
      Benson, Luke  and
      Sun, Lucy  and
      Wardle-Solano, Alexander  and
      Szab{\'o}, Hannah  and
      Zubova, Ekaterina  and
      Burtell, Matthew  and
      Fan, Jonathan  and
      Liu, Yixin  and
      Wong, Brian  and
      Sailor, Malcolm  and
      Ni, Ansong  and
      Nan, Linyong  and
      Kasai, Jungo  and
      Yu, Tao  and
      Zhang, Rui  and
      Fabbri, Alexander  and
      Kryscinski, Wojciech Maciej  and
      Yavuz, Semih  and
      Liu, Ye  and
      Lin, Xi Victoria  and
      Joty, Shafiq  and
      Zhou, Yingbo  and
      Xiong, Caiming  and
      Ying, Rex  and
      Cohan, Arman  and
      Radev, Dragomir",
    editor = "Al-Onaizan, Yaser  and
      Bansal, Mohit  and
      Chen, Yun-Nung",
    booktitle = "Proceedings of the 2024 Conference on Empirical Methods in Natural Language Processing",
    month = nov,
    year = "2024",
    address = "Miami, Florida, USA",
    publisher = "Association for Computational Linguistics",
    url = "https://aclanthology.org/2024.emnlp-main.1229/",
    doi = "10.18653/v1/2024.emnlp-main.1229",
    pages = "22017--22031"
}

@inproceedings{tan-etal-2025-enhancing-logical,
    title = "Enhancing Logical Reasoning in Language Models via Symbolically-Guided {M}onte {C}arlo Process Supervision",
    author = "Tan, Xingwei  and
      Valentino, Marco  and
      Akhter, Mahmud Elahi  and
      Liakata, Maria  and
      Aletras, Nikolaos",
    editor = "Christodoulopoulos, Christos  and
      Chakraborty, Tanmoy  and
      Rose, Carolyn  and
      Peng, Violet",
    booktitle = "Proceedings of the 2025 Conference on Empirical Methods in Natural Language Processing",
    month = nov,
    year = "2025",
    address = "Suzhou, China",
    publisher = "Association for Computational Linguistics",
    url = "https://aclanthology.org/2025.emnlp-main.1624/",
    doi = "10.18653/v1/2025.emnlp-main.1624",
    pages = "31874--31888",
    ISBN = "979-8-89176-332-6"
}

@book{salmon1994introduction,
  title={Introduction to logic and critical thinking},
  author={Salmon, Merrilee H},
  year={1994},
  publisher={Wadsworth Publishing Co Inc},
    ISBN = {978-0155056053}
}

@article{frankfurt1958peirce,
  title={Peirce's notion of abduction},
  author={Frankfurt, Harry G},
  journal={The Journal of Philosophy},
  volume={55},
  number={14},
  pages={593--597},
  year={1958},
  publisher={JSTOR}
}

@article{wielemaker:2011:tplp,
  author = {Wielemaker, Jan and Schrijvers, Tom and Triska, Markus and Lager, Torbj\"o{}rn},
  title = {{SWI-Prolog}},
  journal = {Theory and Practice of Logic Programming},
  year = {2012},
  volume = {12},
  number = {1-2},
  pages = {67--96},
  ISSN = {1471-0684}
}

@inproceedings{haldar-hockenmaier-2025-rating,
    title = "Rating Roulette: Self-Inconsistency in {LLM}-As-A-Judge Frameworks",
    author = "Haldar, Rajarshi  and
      Hockenmaier, Julia",
    editor = "Christodoulopoulos, Christos  and
      Chakraborty, Tanmoy  and
      Rose, Carolyn  and
      Peng, Violet",
    booktitle = "Findings of the Association for Computational Linguistics: EMNLP 2025",
    month = nov,
    year = "2025",
    address = "Suzhou, China",
    publisher = "Association for Computational Linguistics",
    url = "https://aclanthology.org/2025.findings-emnlp.1361/",
    doi = "10.18653/v1/2025.findings-emnlp.1361",
    pages = "24986--25004",
    ISBN = "979-8-89176-335-7"
}

@inproceedings{li-etal-2025-language-models,
    title = "Language Models Can Easily Learn to Reason from Demonstrations",
    author = "Li, Dacheng  and
      Cao, Shiyi  and
      Griggs, Tyler  and
      Liu, Shu  and
      Mo, Xiangxi  and
      Tang, Eric  and
      Hegde, Sumanth  and
      Hakhamaneshi, Kourosh  and
      Patil, Shishir G  and
      Zaharia, Matei  and
      Gonzalez, Joseph E.  and
      Stoica, Ion",
    editor = "Christodoulopoulos, Christos  and
      Chakraborty, Tanmoy  and
      Rose, Carolyn  and
      Peng, Violet",
    booktitle = "Findings of the Association for Computational Linguistics: EMNLP 2025",
    month = nov,
    year = "2025",
    address = "Suzhou, China",
    publisher = "Association for Computational Linguistics",
    url = "https://aclanthology.org/2025.findings-emnlp.866/",
    doi = "10.18653/v1/2025.findings-emnlp.866",
    pages = "15979--15997",
    ISBN = "979-8-89176-335-7"
}

@inproceedings{zheng-etal-2025-logidynamics,
    title = "{L}ogi{D}ynamics: Unraveling the Dynamics of Inductive, Abductive and Deductive Logical Inferences in {LLM} Reasoning",
    author = "Zheng, Tianshi  and
      Jiayang, Cheng  and
      Li, Chunyang  and
      Shi, Haochen  and
      Wang, Zihao  and
      Bai, Jiaxin  and
      Song, Yangqiu  and
      Wong, Ginny  and
      See, Simon",
    editor = "Christodoulopoulos, Christos  and
      Chakraborty, Tanmoy  and
      Rose, Carolyn  and
      Peng, Violet",
    booktitle = "Proceedings of the 2025 Conference on Empirical Methods in Natural Language Processing",
    month = nov,
    year = "2025",
    address = "Suzhou, China",
    publisher = "Association for Computational Linguistics",
    url = "https://aclanthology.org/2025.emnlp-main.1045/",
    doi = "10.18653/v1/2025.emnlp-main.1045",
    pages = "20721--20742",
    ISBN = "979-8-89176-332-6"
}

@misc{olmo2025olmo3,
      title={Olmo 3}, 
      author={Team Olmo and : and Allyson Ettinger and Amanda Bertsch and Bailey Kuehl and David Graham and David Heineman and Dirk Groeneveld and Faeze Brahman and Finbarr Timbers and Hamish Ivison and Jacob Morrison and Jake Poznanski and Kyle Lo and Luca Soldaini and Matt Jordan and Mayee Chen and Michael Noukhovitch and Nathan Lambert and Pete Walsh and Pradeep Dasigi and Robert Berry and Saumya Malik and Saurabh Shah and Scott Geng and Shane Arora and Shashank Gupta and Taira Anderson and Teng Xiao and Tyler Murray and Tyler Romero and Victoria Graf and Akari Asai and Akshita Bhagia and Alexander Wettig and Alisa Liu and Aman Rangapur and Chloe Anastasiades and Costa Huang and Dustin Schwenk and Harsh Trivedi and Ian Magnusson and Jaron Lochner and Jiacheng Liu and Lester James V. Miranda and Maarten Sap and Malia Morgan and Michael Schmitz and Michal Guerquin and Michael Wilson and Regan Huff and Ronan Le Bras and Rui Xin and Rulin Shao and Sam Skjonsberg and Shannon Zejiang Shen and Shuyue Stella Li and Tucker Wilde and Valentina Pyatkin and Will Merrill and Yapei Chang and Yuling Gu and Zhiyuan Zeng and Ashish Sabharwal and Luke Zettlemoyer and Pang Wei Koh and Ali Farhadi and Noah A. Smith and Hannaneh Hajishirzi},
      year={2025},
      eprint={2512.13961},
      archivePrefix={arXiv},
      primaryClass={cs.CL},
      url={https://arxiv.org/abs/2512.13961}, 
}

@inproceedings{dougrez-lewis-etal-2025-assessing,
    title = "Assessing the Reasoning Capabilities of {LLM}s in the context of Evidence-based Claim Verification",
    author = {Dougrez-Lewis, John  and
      Akhter, Mahmud Elahi  and
      Ruggeri, Federico  and
      L{\"o}bbers, Sebastian  and
      He, Yulan  and
      Liakata, Maria},
    editor = "Che, Wanxiang  and
      Nabende, Joyce  and
      Shutova, Ekaterina  and
      Pilehvar, Mohammad Taher",
    booktitle = "Findings of the Association for Computational Linguistics: ACL 2025",
    month = jul,
    year = "2025",
    address = "Vienna, Austria",
    publisher = "Association for Computational Linguistics",
    url = "https://aclanthology.org/2025.findings-acl.1059/",
    doi = "10.18653/v1/2025.findings-acl.1059",
    pages = "20604--20628",
    ISBN = "979-8-89176-256-5"
}

@misc{deepseekai2025deepseekr1incentivizingreasoningcapability,
      title={DeepSeek-R1: Incentivizing Reasoning Capability in LLMs via Reinforcement Learning}, 
      author={DeepSeek-AI},
      year={2025},
      eprint={2501.12948},
      archivePrefix={arXiv},
      primaryClass={cs.CL},
      url={https://arxiv.org/abs/2501.12948}, 
}

@inproceedings{
hu2022lora,
title={Lo{RA}: Low-Rank Adaptation of Large Language Models},
author={Edward J Hu and yelong shen and Phillip Wallis and Zeyuan Allen-Zhu and Yuanzhi Li and Shean Wang and Lu Wang and Weizhu Chen},
booktitle={International Conference on Learning Representations},
year={2022},
url={https://openreview.net/forum?id=nZeVKeeFYf9}
}

@inproceedings{kwon2023efficient,
  title={Efficient Memory Management for Large Language Model Serving with PagedAttention},
  author={Woosuk Kwon and Zhuohan Li and Siyuan Zhuang and Ying Sheng and Lianmin Zheng and Cody Hao Yu and Joseph E. Gonzalez and Hao Zhang and Ion Stoica},
  booktitle={Proceedings of the ACM SIGOPS 29th Symposium on Operating Systems Principles},
  year={2023}
}

@inproceedings{zhao-etal-2023-abductive,
    title = "Abductive Commonsense Reasoning Exploiting Mutually Exclusive Explanations",
    author = "Zhao, Wenting  and
      Chiu, Justin  and
      Cardie, Claire  and
      Rush, Alexander",
    editor = "Rogers, Anna  and
      Boyd-Graber, Jordan  and
      Okazaki, Naoaki",
    booktitle = "Proceedings of the 61st Annual Meeting of the Association for Computational Linguistics (Volume 1: Long Papers)",
    month = jul,
    year = "2023",
    address = "Toronto, Canada",
    publisher = "Association for Computational Linguistics",
    url = "https://aclanthology.org/2023.acl-long.831/",
    doi = "10.18653/v1/2023.acl-long.831",
    pages = "14883--14896"
}

@inproceedings{del-fishel-2023-true,
    title = "True Detective: A Deep Abductive Reasoning Benchmark Undoable for {GPT}-3 and Challenging for {GPT}-4",
    author = "Del, Maksym  and
      Fishel, Mark",
    editor = "Palmer, Alexis  and
      Camacho-collados, Jose",
    booktitle = "Proceedings of the 12th Joint Conference on Lexical and Computational Semantics (*SEM 2023)",
    month = jul,
    year = "2023",
    address = "Toronto, Canada",
    publisher = "Association for Computational Linguistics",
    url = "https://aclanthology.org/2023.starsem-1.28",
    doi = "10.18653/v1/2023.starsem-1.28",
    pages = "314--322",
}

@inproceedings{wu-etal-2024-llama,
    title = "{LL}a{MA} Pro: Progressive {LL}a{MA} with Block Expansion",
    author = "Wu, Chengyue  and
      Gan, Yukang  and
      Ge, Yixiao  and
      Lu, Zeyu  and
      Wang, Jiahao  and
      Feng, Ye  and
      Shan, Ying  and
      Luo, Ping",
    editor = "Ku, Lun-Wei  and
      Martins, Andre  and
      Srikumar, Vivek",
    booktitle = "Proceedings of the 62nd Annual Meeting of the Association for Computational Linguistics (Volume 1: Long Papers)",
    month = aug,
    year = "2024",
    address = "Bangkok, Thailand",
    publisher = "Association for Computational Linguistics",
    url = "https://aclanthology.org/2024.acl-long.352/",
    doi = "10.18653/v1/2024.acl-long.352",
    pages = "6518--6537"
}

@misc{cao2025progressivedepthupscalingoptimal,
      title={Progressive Depth Up-scaling via Optimal Transport}, 
      author={Mingzi Cao and Xi Wang and Nikolaos Aletras},
      year={2025},
      eprint={2508.08011},
      archivePrefix={arXiv},
      primaryClass={cs.CL},
      url={https://arxiv.org/abs/2508.08011}, 
}

@misc{choudhury2025llama3nanda10bchatopengenerativelarge,
      title={Llama-3-Nanda-10B-Chat: An Open Generative Large Language Model for Hindi}, 
      author={Monojit Choudhury and Shivam Chauhan and Rocktim Jyoti Das and Dhruv Sahnan and Xudong Han and Haonan Li and Aaryamonvikram Singh and Alok Anil Jadhav and Utkarsh Agarwal and Mukund Choudhary and Debopriyo Banerjee and Fajri Koto and Junaid Bhat and Awantika Shukla and Samujjwal Ghosh and Samta Kamboj and Onkar Pandit and Lalit Pradhan and Rahul Pal and Sunil Sahu and Soundar Doraiswamy and Parvez Mullah and Ali El Filali and Neha Sengupta and Gokul Ramakrishnan and Rituraj Joshi and Gurpreet Gosal and Avraham Sheinin and Natalia Vassilieva and Preslav Nakov},
      year={2025},
      eprint={2504.06011},
      archivePrefix={arXiv},
      primaryClass={cs.CL},
      url={https://arxiv.org/abs/2504.06011}, 
}

@misc{yang2025qwen3technicalreport,
      title={Qwen3 Technical Report}, 
      author={An Yang and Anfeng Li and Baosong Yang and Beichen Zhang and Binyuan Hui and Bo Zheng and Bowen Yu and Chang Gao and Chengen Huang and Chenxu Lv and Chujie Zheng and Dayiheng Liu and Fan Zhou and Fei Huang and Feng Hu and Hao Ge and Haoran Wei and Huan Lin and Jialong Tang and Jian Yang and Jianhong Tu and Jianwei Zhang and Jianxin Yang and Jiaxi Yang and Jing Zhou and Jingren Zhou and Junyang Lin and Kai Dang and Keqin Bao and Kexin Yang and Le Yu and Lianghao Deng and Mei Li and Mingfeng Xue and Mingze Li and Pei Zhang and Peng Wang and Qin Zhu and Rui Men and Ruize Gao and Shixuan Liu and Shuang Luo and Tianhao Li and Tianyi Tang and Wenbiao Yin and Xingzhang Ren and Xinyu Wang and Xinyu Zhang and Xuancheng Ren and Yang Fan and Yang Su and Yichang Zhang and Yinger Zhang and Yu Wan and Yuqiong Liu and Zekun Wang and Zeyu Cui and Zhenru Zhang and Zhipeng Zhou and Zihan Qiu},
      year={2025},
      eprint={2505.09388},
      archivePrefix={arXiv},
      primaryClass={cs.CL},
      url={https://arxiv.org/abs/2505.09388}, 
}

@misc{grattafiori2024llama3herdmodels,
      title={The Llama 3 Herd of Models}, 
      author={Aaron Grattafiori and Abhimanyu Dubey and Abhinav Jauhri and Abhinav Pandey and Abhishek Kadian and Ahmad Al-Dahle and Aiesha Letman and Akhil Mathur and Alan Schelten and Alex Vaughan and Amy Yang and Angela Fan and Anirudh Goyal and Anthony Hartshorn and Aobo Yang and Archi Mitra and Archie Sravankumar and Artem Korenev and Arthur Hinsvark and Arun Rao and Aston Zhang and Aurelien Rodriguez and Austen Gregerson and Ava Spataru and Baptiste Roziere and Bethany Biron and Binh Tang and Bobbie Chern and Charlotte Caucheteux and Chaya Nayak and Chloe Bi and Chris Marra and Chris McConnell and Christian Keller and Christophe Touret and Chunyang Wu and Corinne Wong and Cristian Canton Ferrer and Cyrus Nikolaidis and Damien Allonsius and Daniel Song and Danielle Pintz and Danny Livshits and Danny Wyatt and David Esiobu and Dhruv Choudhary and Dhruv Mahajan and Diego Garcia-Olano and Diego Perino and Dieuwke Hupkes and Egor Lakomkin and Ehab AlBadawy and Elina Lobanova and Emily Dinan and Eric Michael Smith and Filip Radenovic and Francisco Guzmán and Frank Zhang and Gabriel Synnaeve and Gabrielle Lee and Georgia Lewis Anderson and Govind Thattai and Graeme Nail and Gregoire Mialon and Guan Pang and Guillem Cucurell and Hailey Nguyen and Hannah Korevaar and Hu Xu and Hugo Touvron and Iliyan Zarov and Imanol Arrieta Ibarra and Isabel Kloumann and Ishan Misra and Ivan Evtimov and Jack Zhang and Jade Copet and Jaewon Lee and Jan Geffert and Jana Vranes and Jason Park and Jay Mahadeokar and Jeet Shah and Jelmer van der Linde and Jennifer Billock and Jenny Hong and Jenya Lee and Jeremy Fu and Jianfeng Chi and Jianyu Huang and Jiawen Liu and Jie Wang and Jiecao Yu and Joanna Bitton and Joe Spisak and Jongsoo Park and Joseph Rocca and Joshua Johnstun and Joshua Saxe and Junteng Jia and Kalyan Vasuden Alwala and Karthik Prasad and Kartikeya Upasani and Kate Plawiak and Ke Li and Kenneth Heafield and Kevin Stone and Khalid El-Arini and Krithika Iyer and Kshitiz Malik and Kuenley Chiu and Kunal Bhalla and Kushal Lakhotia and Lauren Rantala-Yeary and Laurens van der Maaten and Lawrence Chen and Liang Tan and Liz Jenkins and Louis Martin and Lovish Madaan and Lubo Malo and Lukas Blecher and Lukas Landzaat and Luke de Oliveira and Madeline Muzzi and Mahesh Pasupuleti and Mannat Singh and Manohar Paluri and Marcin Kardas and Maria Tsimpoukelli and Mathew Oldham and Mathieu Rita and Maya Pavlova and Melanie Kambadur and Mike Lewis and Min Si and Mitesh Kumar Singh and Mona Hassan and Naman Goyal and Narjes Torabi and Nikolay Bashlykov and Nikolay Bogoychev and Niladri Chatterji and Ning Zhang and Olivier Duchenne and Onur Çelebi and Patrick Alrassy and Pengchuan Zhang and Pengwei Li and Petar Vasic and Peter Weng and Prajjwal Bhargava and Pratik Dubal and Praveen Krishnan and Punit Singh Koura and Puxin Xu and Qing He and Qingxiao Dong and Ragavan Srinivasan and Raj Ganapathy and Ramon Calderer and Ricardo Silveira Cabral and Robert Stojnic and Roberta Raileanu and Rohan Maheswari and Rohit Girdhar and Rohit Patel and Romain Sauvestre and Ronnie Polidoro and Roshan Sumbaly and Ross Taylor and Ruan Silva and Rui Hou and Rui Wang and Saghar Hosseini and Sahana Chennabasappa and Sanjay Singh and Sean Bell and Seohyun Sonia Kim and Sergey Edunov and Shaoliang Nie and Sharan Narang and Sharath Raparthy and Sheng Shen and Shengye Wan and Shruti Bhosale and Shun Zhang and Simon Vandenhende and Soumya Batra and Spencer Whitman and Sten Sootla and Stephane Collot and Suchin Gururangan and Sydney Borodinsky and Tamar Herman and Tara Fowler and Tarek Sheasha and Thomas Georgiou and Thomas Scialom and Tobias Speckbacher and Todor Mihaylov and Tong Xiao and Ujjwal Karn and Vedanuj Goswami and Vibhor Gupta and Vignesh Ramanathan and Viktor Kerkez and Vincent Gonguet and Virginie Do and Vish Vogeti and Vítor Albiero and Vladan Petrovic and Weiwei Chu and Wenhan Xiong and Wenyin Fu and Whitney Meers and Xavier Martinet and Xiaodong Wang and Xiaofang Wang and Xiaoqing Ellen Tan and Xide Xia and Xinfeng Xie and Xuchao Jia and Xuewei Wang and Yaelle Goldschlag and Yashesh Gaur and Yasmine Babaei and Yi Wen and Yiwen Song and Yuchen Zhang and Yue Li and Yuning Mao and Zacharie Delpierre Coudert and Zheng Yan and Zhengxing Chen and Zoe Papakipos and Aaditya Singh and Aayushi Srivastava and Abha Jain and Adam Kelsey and Adam Shajnfeld and Adithya Gangidi and Adolfo Victoria and Ahuva Goldstand and Ajay Menon and Ajay Sharma and Alex Boesenberg and Alexei Baevski and Allie Feinstein and Amanda Kallet and Amit Sangani and Amos Teo and Anam Yunus and Andrei Lupu and Andres Alvarado and Andrew Caples and Andrew Gu and Andrew Ho and Andrew Poulton and Andrew Ryan and Ankit Ramchandani and Annie Dong and Annie Franco and Anuj Goyal and Aparajita Saraf and Arkabandhu Chowdhury and Ashley Gabriel and Ashwin Bharambe and Assaf Eisenman and Azadeh Yazdan and Beau James and Ben Maurer and Benjamin Leonhardi and Bernie Huang and Beth Loyd and Beto De Paola and Bhargavi Paranjape and Bing Liu and Bo Wu and Boyu Ni and Braden Hancock and Bram Wasti and Brandon Spence and Brani Stojkovic and Brian Gamido and Britt Montalvo and Carl Parker and Carly Burton and Catalina Mejia and Ce Liu and Changhan Wang and Changkyu Kim and Chao Zhou and Chester Hu and Ching-Hsiang Chu and Chris Cai and Chris Tindal and Christoph Feichtenhofer and Cynthia Gao and Damon Civin and Dana Beaty and Daniel Kreymer and Daniel Li and David Adkins and David Xu and Davide Testuggine and Delia David and Devi Parikh and Diana Liskovich and Didem Foss and Dingkang Wang and Duc Le and Dustin Holland and Edward Dowling and Eissa Jamil and Elaine Montgomery and Eleonora Presani and Emily Hahn and Emily Wood and Eric-Tuan Le and Erik Brinkman and Esteban Arcaute and Evan Dunbar and Evan Smothers and Fei Sun and Felix Kreuk and Feng Tian and Filippos Kokkinos and Firat Ozgenel and Francesco Caggioni and Frank Kanayet and Frank Seide and Gabriela Medina Florez and Gabriella Schwarz and Gada Badeer and Georgia Swee and Gil Halpern and Grant Herman and Grigory Sizov and Guangyi and Zhang and Guna Lakshminarayanan and Hakan Inan and Hamid Shojanazeri and Han Zou and Hannah Wang and Hanwen Zha and Haroun Habeeb and Harrison Rudolph and Helen Suk and Henry Aspegren and Hunter Goldman and Hongyuan Zhan and Ibrahim Damlaj and Igor Molybog and Igor Tufanov and Ilias Leontiadis and Irina-Elena Veliche and Itai Gat and Jake Weissman and James Geboski and James Kohli and Janice Lam and Japhet Asher and Jean-Baptiste Gaya and Jeff Marcus and Jeff Tang and Jennifer Chan and Jenny Zhen and Jeremy Reizenstein and Jeremy Teboul and Jessica Zhong and Jian Jin and Jingyi Yang and Joe Cummings and Jon Carvill and Jon Shepard and Jonathan McPhie and Jonathan Torres and Josh Ginsburg and Junjie Wang and Kai Wu and Kam Hou U and Karan Saxena and Kartikay Khandelwal and Katayoun Zand and Kathy Matosich and Kaushik Veeraraghavan and Kelly Michelena and Keqian Li and Kiran Jagadeesh and Kun Huang and Kunal Chawla and Kyle Huang and Lailin Chen and Lakshya Garg and Lavender A and Leandro Silva and Lee Bell and Lei Zhang and Liangpeng Guo and Licheng Yu and Liron Moshkovich and Luca Wehrstedt and Madian Khabsa and Manav Avalani and Manish Bhatt and Martynas Mankus and Matan Hasson and Matthew Lennie and Matthias Reso and Maxim Groshev and Maxim Naumov and Maya Lathi and Meghan Keneally and Miao Liu and Michael L. Seltzer and Michal Valko and Michelle Restrepo and Mihir Patel and Mik Vyatskov and Mikayel Samvelyan and Mike Clark and Mike Macey and Mike Wang and Miquel Jubert Hermoso and Mo Metanat and Mohammad Rastegari and Munish Bansal and Nandhini Santhanam and Natascha Parks and Natasha White and Navyata Bawa and Nayan Singhal and Nick Egebo and Nicolas Usunier and Nikhil Mehta and Nikolay Pavlovich Laptev and Ning Dong and Norman Cheng and Oleg Chernoguz and Olivia Hart and Omkar Salpekar and Ozlem Kalinli and Parkin Kent and Parth Parekh and Paul Saab and Pavan Balaji and Pedro Rittner and Philip Bontrager and Pierre Roux and Piotr Dollar and Polina Zvyagina and Prashant Ratanchandani and Pritish Yuvraj and Qian Liang and Rachad Alao and Rachel Rodriguez and Rafi Ayub and Raghotham Murthy and Raghu Nayani and Rahul Mitra and Rangaprabhu Parthasarathy and Raymond Li and Rebekkah Hogan and Robin Battey and Rocky Wang and Russ Howes and Ruty Rinott and Sachin Mehta and Sachin Siby and Sai Jayesh Bondu and Samyak Datta and Sara Chugh and Sara Hunt and Sargun Dhillon and Sasha Sidorov and Satadru Pan and Saurabh Mahajan and Saurabh Verma and Seiji Yamamoto and Sharadh Ramaswamy and Shaun Lindsay and Shaun Lindsay and Sheng Feng and Shenghao Lin and Shengxin Cindy Zha and Shishir Patil and Shiva Shankar and Shuqiang Zhang and Shuqiang Zhang and Sinong Wang and Sneha Agarwal and Soji Sajuyigbe and Soumith Chintala and Stephanie Max and Stephen Chen and Steve Kehoe and Steve Satterfield and Sudarshan Govindaprasad and Sumit Gupta and Summer Deng and Sungmin Cho and Sunny Virk and Suraj Subramanian and Sy Choudhury and Sydney Goldman and Tal Remez and Tamar Glaser and Tamara Best and Thilo Koehler and Thomas Robinson and Tianhe Li and Tianjun Zhang and Tim Matthews and Timothy Chou and Tzook Shaked and Varun Vontimitta and Victoria Ajayi and Victoria Montanez and Vijai Mohan and Vinay Satish Kumar and Vishal Mangla and Vlad Ionescu and Vlad Poenaru and Vlad Tiberiu Mihailescu and Vladimir Ivanov and Wei Li and Wenchen Wang and Wenwen Jiang and Wes Bouaziz and Will Constable and Xiaocheng Tang and Xiaojian Wu and Xiaolan Wang and Xilun Wu and Xinbo Gao and Yaniv Kleinman and Yanjun Chen and Ye Hu and Ye Jia and Ye Qi and Yenda Li and Yilin Zhang and Ying Zhang and Yossi Adi and Youngjin Nam and Yu and Wang and Yu Zhao and Yuchen Hao and Yundi Qian and Yunlu Li and Yuzi He and Zach Rait and Zachary DeVito and Zef Rosnbrick and Zhaoduo Wen and Zhenyu Yang and Zhiwei Zhao and Zhiyu Ma},
      year={2024},
      eprint={2407.21783},
      archivePrefix={arXiv},
      primaryClass={cs.AI},
      url={https://arxiv.org/abs/2407.21783}, 
}

@misc{hoffmann2025llamagenba10btrilinguallargelanguage,
      title={Llama-GENBA-10B: A Trilingual Large Language Model for German, English and Bavarian}, 
      author={Michael Hoffmann and Jophin John and Stefan Schweter and Gokul Ramakrishnan and Hoi-Fong Mak and Alice Zhang and Dmitry Gaynullin and Nicolay J. Hammer},
      year={2025},
      eprint={2509.05668},
      archivePrefix={arXiv},
      primaryClass={cs.CL},
      url={https://arxiv.org/abs/2509.05668}, 
}

@inproceedings{
nakamura2025dropupcycling,
title={Drop-Upcycling: Training Sparse Mixture of Experts with Partial Re-initialization},
author={Taishi Nakamura and Takuya Akiba and Kazuki Fujii and Yusuke Oda and Rio Yokota and Jun Suzuki},
booktitle={The Thirteenth International Conference on Learning Representations},
year={2025},
url={https://openreview.net/forum?id=gx1wHnf5Vp}
}

@inproceedings{lee-etal-2025-enginius,
    title = "{ENG}inius: A Bilingual {LLM} Optimized for Plant Construction Engineering",
    author = "Lee, Wooseong  and
      Kim, Minseo  and
      Hur, Taeil  and
      Jang, Gyeong Hwan  and
      Lee, Woncheol  and
      Na, Maro  and
      Kim, Taeuk",
    editor = "Rehm, Georg  and
      Li, Yunyao",
    booktitle = "Proceedings of the 63rd Annual Meeting of the Association for Computational Linguistics (Volume 6: Industry Track)",
    month = jul,
    year = "2025",
    address = "Vienna, Austria",
    publisher = "Association for Computational Linguistics",
    url = "https://aclanthology.org/2025.acl-industry.95/",
    doi = "10.18653/v1/2025.acl-industry.95",
    pages = "1350--1364",
    ISBN = "979-8-89176-288-6"
}

@misc{lim2025expandingfoundationallanguagecapabilities,
      title={Expanding Foundational Language Capabilities in Open-Source LLMs through a Korean Case Study}, 
      author={Junghwan Lim and Gangwon Jo and Sungmin Lee and Jiyoung Park and Dongseok Kim and Jihwan Kim and Junhyeok Lee and Wai Ting Cheung and Dahye Choi and Kibong Choi and Jaeyeon Huh and Beomgyu Kim and Jangwoong Kim and Taehyun Kim and Haesol Lee and Jeesoo Lee and Dongpin Oh and Changseok Song and Daewon Suh},
      year={2025},
      eprint={2509.03972},
      archivePrefix={arXiv},
      primaryClass={cs.CL},
      url={https://arxiv.org/abs/2509.03972}, 
}

@inproceedings{
komatsuzaki2023sparse,
title={Sparse Upcycling: Training Mixture-of-Experts from Dense Checkpoints},
author={Aran Komatsuzaki and Joan Puigcerver and James Lee-Thorp and Carlos Riquelme Ruiz and Basil Mustafa and Joshua Ainslie and Yi Tay and Mostafa Dehghani and Neil Houlsby},
booktitle={The Eleventh International Conference on Learning Representations },
year={2023},
url={https://openreview.net/forum?id=T5nUQDrM4u}
}

@misc{vonwerra2022trl,
  author = {Leandro von Werra and Younes Belkada and Lewis Tunstall and Edward Beeching and Tristan Thrush and Nathan Lambert and Shengyi Huang and Kashif Rasul and Quentin Gallouédec},
  title = {TRL: Transformer Reinforcement Learning},
  year = {2020},
  publisher = {GitHub},
  journal = {GitHub repository},
  howpublished = {\url{https://github.com/huggingface/trl}}
}

@inproceedings{dao2023flashattention2,
  title={Flash{A}ttention-2: Faster Attention with Better Parallelism and Work Partitioning},
  author={Dao, Tri},
  booktitle={International Conference on Learning Representations (ICLR)},
  year={2024}
}

@inproceedings{pletenev-etal-2025-much,
    title = "How Much Knowledge Can You Pack into a {L}o{RA} Adapter without Harming {LLM}?",
    author = "Pletenev, Sergey  and
      Marina, Maria  and
      Moskovskiy, Daniil  and
      Konovalov, Vasily  and
      Braslavski, Pavel  and
      Panchenko, Alexander  and
      Salnikov, Mikhail",
    editor = "Chiruzzo, Luis  and
      Ritter, Alan  and
      Wang, Lu",
    booktitle = "Findings of the Association for Computational Linguistics: NAACL 2025",
    month = apr,
    year = "2025",
    address = "Albuquerque, New Mexico",
    publisher = "Association for Computational Linguistics",
    url = "https://aclanthology.org/2025.findings-naacl.243/",
    doi = "10.18653/v1/2025.findings-naacl.243",
    pages = "4309--4322",
    ISBN = "979-8-89176-195-7"
}

@misc{yu2025finemedlmo1enhancingmedicalknowledge,
      title={FineMedLM-o1: Enhancing Medical Knowledge Reasoning Ability of LLM from Supervised Fine-Tuning to Test-Time Training}, 
      author={Hongzhou Yu and Tianhao Cheng and Yingwen Wang and Wen He and Qing Wang and Ying Cheng and Yuejie Zhang and Rui Feng and Xiaobo Zhang},
      year={2025},
      eprint={2501.09213},
      archivePrefix={arXiv},
      primaryClass={cs.CL},
      url={https://arxiv.org/abs/2501.09213}, 
}

@article{
biderman2024lora,
title={Lo{RA} Learns Less and Forgets Less},
author={Dan Biderman and Jacob Portes and Jose Javier Gonzalez Ortiz and Mansheej Paul and Philip Greengard and Connor Jennings and Daniel King and Sam Havens and Vitaliy Chiley and Jonathan Frankle and Cody Blakeney and John Patrick Cunningham},
journal={Transactions on Machine Learning Research},
issn={2835-8856},
year={2024},
url={https://openreview.net/forum?id=aloEru2qCG},
note={Featured Certification}
}

@article{JMLR:v23:21-0998,
  author  = {William Fedus and Barret Zoph and Noam Shazeer},
  title   = {Switch Transformers: Scaling to Trillion Parameter Models with Simple and Efficient Sparsity},
  journal = {Journal of Machine Learning Research},
  year    = {2022},
  volume  = {23},
  number  = {120},
  pages   = {1--39},
  url     = {http://jmlr.org/papers/v23/21-0998.html}
}

@inproceedings{10.1609/aaai.v39i24.34805,
author = {Zhou, Hao and Wang, Zhijun and Huang, Shujian and Huang, Xin and Han, Xue and Feng, Junlan and Deng, Chao and Luo, Weihua and Chen, Jiajun},
title = {MoE-LPR: multilingual extension of large language models through mixture-of-experts with language priors routing},
year = {2025},
isbn = {978-1-57735-897-8},
publisher = {AAAI Press},
url = {https://doi.org/10.1609/aaai.v39i24.34805},
doi = {10.1609/aaai.v39i24.34805},
booktitle = {Proceedings of the Thirty-Ninth AAAI Conference on Artificial Intelligence and Thirty-Seventh Conference on Innovative Applications of Artificial Intelligence and Fifteenth Symposium on Educational Advances in Artificial Intelligence},
articleno = {2907},
numpages = {9},
series = {AAAI'25/IAAI'25/EAAI'25}
}

@inproceedings{zhang-etal-2025-less,
    title = "Less, but Better: Efficient Multilingual Expansion for {LLM}s via Layer-wise Mixture-of-Experts",
    author = "Zhang, Xue  and
      Liang, Yunlong  and
      Meng, Fandong  and
      Zhang, Songming  and
      Chen, Yufeng  and
      Xu, Jinan  and
      Zhou, Jie",
    editor = "Che, Wanxiang  and
      Nabende, Joyce  and
      Shutova, Ekaterina  and
      Pilehvar, Mohammad Taher",
    booktitle = "Proceedings of the 63rd Annual Meeting of the Association for Computational Linguistics (Volume 1: Long Papers)",
    month = jul,
    year = "2025",
    address = "Vienna, Austria",
    publisher = "Association for Computational Linguistics",
    url = "https://aclanthology.org/2025.acl-long.878/",
    doi = "10.18653/v1/2025.acl-long.878",
    pages = "17948--17963",
    ISBN = "979-8-89176-251-0"
}

@misc{aggarwal2025bayesiangeometrytransformerattention,
      title={The Bayesian Geometry of Transformer Attention}, 
      author={Naman Aggarwal and Siddhartha R. Dalal and Vishal Misra},
      year={2025},
      eprint={2512.22471},
      archivePrefix={arXiv},
      primaryClass={cs.LG},
      url={https://arxiv.org/abs/2512.22471}, 
}

@inproceedings{li2025dynamicexpertspecializationcatastrophic,
    title = "Dynamic Expert Specialization: Towards Catastrophic Forgetting-Free Multi-Domain {M}o{E} Adaptation",
    author = "Li, Junzhuo  and
      Wang, Bo  and
      Zhou, Xiuze  and
      Hu, Xuming",
    editor = "Christodoulopoulos, Christos  and
      Chakraborty, Tanmoy  and
      Rose, Carolyn  and
      Peng, Violet",
    booktitle = "Proceedings of the 2025 Conference on Empirical Methods in Natural Language Processing",
    month = nov,
    year = "2025",
    address = "Suzhou, China",
    publisher = "Association for Computational Linguistics",
    url = "https://aclanthology.org/2025.emnlp-main.932/",
    doi = "10.18653/v1/2025.emnlp-main.932",
    pages = "18478--18493",
    ISBN = "979-8-89176-332-6"
}

@misc{wen2025reasoningscaffoldingdistillingflow,
      title={Reasoning Scaffolding: Distilling the Flow of Thought from LLMs}, 
      author={Xiangyu Wen and Junhua Huang and Zeju Li and Min Li and Jianyuan Zhong and Zhijian Xu and Mingxuan Yuan and Yongxiang Huang and Qiang Xu},
      year={2025},
      eprint={2509.23619},
      archivePrefix={arXiv},
      primaryClass={cs.AI},
      url={https://arxiv.org/abs/2509.23619}, 
}

@misc{liu2024datasetslargelanguagemodels,
      title={Datasets for Large Language Models: A Comprehensive Survey}, 
      author={Yang Liu and Jiahuan Cao and Chongyu Liu and Kai Ding and Lianwen Jin},
      year={2024},
      eprint={2402.18041},
      archivePrefix={arXiv},
      primaryClass={cs.CL},
      url={https://arxiv.org/abs/2402.18041}, 
}

@inproceedings{NEURIPS2024_8678da90,
 author = {Morishita, Terufumi and Morio, Gaku and Yamaguchi, Atsuki and Sogawa, Yasuhiro},
 booktitle = {Advances in Neural Information Processing Systems},
 doi = {10.52202/079017-2340},
 editor = {A. Globerson and L. Mackey and D. Belgrave and A. Fan and U. Paquet and J. Tomczak and C. Zhang},
 pages = {73572--73604},
 publisher = {Curran Associates, Inc.},
 title = {Enhancing Reasoning Capabilities of LLMs via Principled Synthetic Logic Corpus},
 url = {https://proceedings.neurips.cc/paper_files/paper/2024/file/8678da90126aa58326b2fc0254b33a8c-Paper-Conference.pdf},
 volume = {37},
 year = {2024}
}

@inproceedings{sun-etal-2024-rapidly,
    title = "Rapidly Developing High-quality Instruction Data and Evaluation Benchmark for Large Language Models with Minimal Human Effort: A Case Study on {J}apanese",
    author = "Sun, Yikun  and
      Wan, Zhen  and
      Ueda, Nobuhiro  and
      Yahata, Sakiko  and
      Cheng, Fei  and
      Chu, Chenhui  and
      Kurohashi, Sadao",
    editor = "Calzolari, Nicoletta  and
      Kan, Min-Yen  and
      Hoste, Veronique  and
      Lenci, Alessandro  and
      Sakti, Sakriani  and
      Xue, Nianwen",
    booktitle = "Proceedings of the 2024 Joint International Conference on Computational Linguistics, Language Resources and Evaluation (LREC-COLING 2024)",
    month = may,
    year = "2024",
    address = "Torino, Italia",
    publisher = "ELRA and ICCL",
    url = "https://aclanthology.org/2024.lrec-main.1184/",
    pages = "13537--13547"
}

@inproceedings{
yu2024metamath,
title={MetaMath: Bootstrap Your Own Mathematical Questions for Large Language Models},
author={Longhui Yu and Weisen Jiang and Han Shi and Jincheng YU and Zhengying Liu and Yu Zhang and James Kwok and Zhenguo Li and Adrian Weller and Weiyang Liu},
booktitle={The Twelfth International Conference on Learning Representations},
year={2024},
url={https://openreview.net/forum?id=N8N0hgNDRt}
}

@misc{zhao2025agentardeepfinance100klargescalefinancialdataset,
      title={Agentar-DeepFinance-100K: A Large-Scale Financial Dataset via Systematic Chain-of-Thought Synthesis Optimization}, 
      author={Xiaoke Zhao and Zhaowen Zhou and Lin Chen and Lihong Wang and Zhiyi Huang and Kaiyuan Zheng and Yanjun Zheng and Xiyang Du and Longfei Liao and Jiawei Liu and Xiang Qi and Bo Zhang and Peng Zhang and Wei Wang and Zhe Li},
      year={2025},
      eprint={2507.12901},
      archivePrefix={arXiv},
      primaryClass={cs.CE},
      url={https://arxiv.org/abs/2507.12901}, 
}

@misc{guha2025openthoughtsdatarecipesreasoning,
      title={OpenThoughts: Data Recipes for Reasoning Models}, 
      author={Etash Guha and Ryan Marten and Sedrick Keh and Negin Raoof and Georgios Smyrnis and Hritik Bansal and Marianna Nezhurina and Jean Mercat and Trung Vu and Zayne Sprague and Ashima Suvarna and Benjamin Feuer and Liangyu Chen and Zaid Khan and Eric Frankel and Sachin Grover and Caroline Choi and Niklas Muennighoff and Shiye Su and Wanjia Zhao and John Yang and Shreyas Pimpalgaonkar and Kartik Sharma and Charlie Cheng-Jie Ji and Yichuan Deng and Sarah Pratt and Vivek Ramanujan and Jon Saad-Falcon and Jeffrey Li and Achal Dave and Alon Albalak and Kushal Arora and Blake Wulfe and Chinmay Hegde and Greg Durrett and Sewoong Oh and Mohit Bansal and Saadia Gabriel and Aditya Grover and Kai-Wei Chang and Vaishaal Shankar and Aaron Gokaslan and Mike A. Merrill and Tatsunori Hashimoto and Yejin Choi and Jenia Jitsev and Reinhard Heckel and Maheswaran Sathiamoorthy and Alexandros G. Dimakis and Ludwig Schmidt},
      year={2025},
      eprint={2506.04178},
      archivePrefix={arXiv},
      primaryClass={cs.LG},
      url={https://arxiv.org/abs/2506.04178}, 
}

@inproceedings{gritsch-etal-2025-nexus,
    title = "Nexus: Adaptive Upcycling to Efficiently Pretrain Mixture of Experts",
    author = {Gritsch, Nikolas  and
      Zhang, Qizhen  and
      Locatelli, Acyr  and
      Hooker, Sara  and
      {\"U}st{\"u}n, Ahmet},
    editor = "Christodoulopoulos, Christos  and
      Chakraborty, Tanmoy  and
      Rose, Carolyn  and
      Peng, Violet",
    booktitle = "Findings of the Association for Computational Linguistics: EMNLP 2025",
    month = nov,
    year = "2025",
    address = "Suzhou, China",
    publisher = "Association for Computational Linguistics",
    url = "https://aclanthology.org/2025.findings-emnlp.1323/",
    doi = "10.18653/v1/2025.findings-emnlp.1323",
    pages = "24364--24381",
    ISBN = "979-8-89176-335-7"
}

@inproceedings{10.5555/3031843.3031909,
author = {Levesque, Hector J. and Davis, Ernest and Morgenstern, Leora},
title = {The Winograd schema challenge},
year = {2012},
isbn = {9781577355601},
publisher = {AAAI Press},
booktitle = {Proceedings of the Thirteenth International Conference on Principles of Knowledge Representation and Reasoning},
pages = {552–561},
numpages = {10},
location = {Rome, Italy},
series = {KR'12}
}

\clearpage
\appendix

\section{Reasoning Task Examples}
\label{app:data_examples}

The following boxes present the questions and corresponding gold answers used in our synthetic fundamental reasoning tasks.

\begin{tcolorbox}[float=htb, colback=green!5, colframe=green!60!black, title=Deductive Task]
\scriptsize
This is a <Deductive> reasoning task. Below are some formulas connected by conjunctions:\\
 $\neg(((\neg A \land A) \lor \neg(F)))$\\
 $\quad \land  ((\neg(\neg E) \lor \neg (\neg B)) \oplus ((H \land F) \rightarrow (H \leftrightarrow F)))$\\
 $\land  \neg (((H \rightarrow A) \lor (\neg G \oplus D)))$\\
 $\land  (((F \land C) \land (G \oplus \neg G)) \leftrightarrow ((D \leftrightarrow A) \land (F \land G)))$\\
 $\land  ((\neg (C) \land (\neg F \rightarrow D)) \oplus \neg ((\neg B \leftrightarrow \neg F)))$\\
Please list the truth value of each variable to make the whole conjunction true using a JSON dictionary, which maps variable names to their truth values, then enclose the answer in <answer><answer>. Please put all the intermediate reasoning steps in <think><think>.
\\
Gold Answer: \\
\{
``A'': ``False'',``B'': ``False'',\\
``C'': ``True'',\\
``D'': ``False'',\\
``E'': ``False'',\\
``F'': ``True'',\\
``G'': ``True'',\\
``H'': ``True''\\
\}
\end{tcolorbox}

\begin{tcolorbox}[float=htb, colback=green!5, colframe=blue!60!black, title=Inductive Task]
\scriptsize
This is a <Inductive> reasoning task. Given the following sequence,\\
$[$`5', `10', `6', `12', `15', `30', `26', `52', `55', `110', `?'$]$\\
What is the value at the question mark? Please enclose the answer in <answer><answer>, and put all the intermediate reasoning steps in <think><think>.\\
Gold Answer: 106
\end{tcolorbox}

\begin{table}[htb]
  \centering
  \resizebox{0.95\columnwidth}{!}{%
  \begin{tabularx}{\columnwidth}{|X|X|}
  \hline
  \rowcolor{rcolor}
  \tiny
Rev. Rule-Graph Search \citep{hu2025ahasystematicmetaabilitiesalignment} &
  \tiny
Assumption Trace-Back (Ours)
\\
\hline

\tiny
\textbf{Premises:} [`(L) => L', 

`(((NOT D) OR (NOT M))) => N',

`((M OR M)) => C', 

`((M OR L)) => LI', 

`(M) => M', 

`((L OR LI)) => G']

\textbf{Known Atoms:} [`L', `M', `LI', `D', `N']

\textbf{Goals:} [`LI', `KM', `NK']

\textbf{Instruction:} For each goal, first assume it is true and determine what conditions must be satisfied. Identify which premises directly lead to these goals, then trace back what must be established to satisfy those premises. Finally, check whether the known atoms can lead to these conditions

&
\tiny
\textbf{Premises:} [`(L) => L', 

`(((NOT D) OR (NOT M))) => N',

`((M OR M)) => C', 

`((M OR L)) => B', 

`(M) => M', 

`((L OR B)) => G']

\textbf{Known Atoms:} ['L', 'M', 'A', 'D', 'N']

\textbf{Goals:} ['B', 'A', 'C']

\textbf{Instruction:} For each goal, identify which premises directly lead to the goal. Then, trace back what the true value of the atoms must be to make each of the goal true. Only the atoms in the `known atoms' are known but their values are not shown. Finally, return the reachable goals with the true values of the known atoms that make it true. Please enclose the final answer with <answer><answer>. All the intermediate thinking steps should be enclosed in <think><think> tags.
\\
\hline
\tiny
\textbf{Answer:} (1) B is reachable (2) KM is unreachable (3) NK is unreachable

&
\tiny
\textbf{Answer:} Goal `B' is reachable. 

Sol 1: \{``D'': false, ``L'': false, ``M'': false\}

Sol 2: \{``D'': false, ``L'': false, ``M'': true\}

...

Sol 8: \{``D'': true, ``L'': true, ``M'': true\}
\\ \hline
  \end{tabularx}%
  }
  \caption{Comparison between the original abductive task by \citet{hu2025ahasystematicmetaabilitiesalignment} and ours.}
  \label{tab:abductive_example}
\end{table}

\section{Implementation Details}
\label{sec:imp_detail}

\paragraph{Data Sampling.} The generation of training data takes approximately 280 hours for LLaMA trajectories and 304 hours for Qwen trajectories when executed on an AMD MI300X accelerator.

\paragraph{Training and Evaluation Setup} 
For training all methods, we use a cosine scheduler with warm-up and a maximum learning rate of $5e$-$6$ (degraded to $5e$-$7$). We fine-tune models using SFT trainer from package TRL \citep{vonwerra2022trl} on one NVIDIA A100 (80GB). We employ Flash-Attention 2 \citep{dao2023flashattention2} and mixed-precision \texttt{bf16} for accelerated training. We use the package Transformers \citep{wolf-etal-2020-transformers} for reproducible output generation and vLLM \citep{kwon2023efficient} for evaluation.

\paragraph{Format-Robust Evaluation.}
 Due to requirements (\S~\ref{sec:our_tasks}), the final answer is explicitly separated from the reasoning chain, and Deduction and Abduction tasks need JSON-formatted answers. Induced models sometimes deviate from the required format while still producing correct answers.

To address this issue, we use \textit{Qwen3-30B-Instruct} \citep{yang2025qwen3technicalreport} to compare the generated answer with the gold answer. Specifically, the model is instructed to determine whether the generation is consistent with the gold label and to return \textit{TRUE} or \textit{FALSE}. The full prompt is provided in Appendix~\ref{sec:prompt}.

\section{Prompt Examples}
\label{sec:prompt}

Below is the prompt used to compare generated answers with gold answers.

\begin{tcolorbox}[float=!htb, colback=green!5, colframe=green!60!black, title=Flexible Match]
\scriptsize
Instruction: Please check whether the generation results is consistent with the gold label.\\

Generation Results:\{output\}\\

Gold Label:\{answer\}\\

Please output TRUE if they are consistent, otherwise output FALSE.

\end{tcolorbox}

\section{Full Results}
\label{sec:Full_Results}

Tables~\ref{tab:results} and~\ref{tab:qwen_results} present the full results of inducing reasoning paradigms using different inducing methods across experimental settings.

\begin{table*}[!t]
\begin{center}
\small
\setlength{\tabcolsep}{1mm}
\begin{adjustbox}{width=0.9\textwidth}
\begin{tabular}{lccccccccccc}
\toprule
& & & \multicolumn{3}{c}{\textbf{Symbolic Tasks}} & \multicolumn{6}{c}{\textbf{Realistic Tasks}} \\
\cmidrule(l{2pt}r{2pt}){4-6} \cmidrule(l{2pt}r{2pt}){7-12}
\textbf{Method} & \textbf{FT} & \textbf{Train/Total} & \textbf{Deduct} & \textbf{Induct} & \textbf{Abduct} & \textbf{Detect} & \textbf{$\alpha$NLI} & \textbf{Winowhy} & \textbf{Folio} & \textbf{RECV} & \textbf{Avg} \\
\midrule

\rowcolor{gray!20} Llama-3.3-70B-IT & - & 0/70.0 &
33.00$_{\pm0.3}$ & 55.33$_{\pm2.5}$ & 16.00$_{\pm2.6}$ &
39.96$_{\pm0.3}$ & 79.28$_{\pm0.2}$ & 65.75$_{\pm0.2}$ &
60.59$_{\pm0.9}$ & 77.07$_{\pm0.1}$ & 64.53$_{\pm0.2}$ \\

\rowcolor{gray!20} Llama-3.1-8B-IT & - & 0/8.03 &
14.00$_{\pm1.0}$ & 32.00$_{\pm0.0}$ & 1.67$_{\pm0.6}$ & 28.80$_{\pm0.0}$ &
73.97$_{\pm0.0}$ & 48.60$_{\pm0.4}$ & 53.69$_{\pm0.5}$ & 71.98$_{\pm0.0}$ & 55.41$_{\pm0.0}$ \\

\midrule
Full FT & & 8.03/8.03 &
\colorbox[HTML]{67FD9A}{\textbf{15.00$_{\pm1.0}$}} & \colorbox[HTML]{67FD9A}{\underline{57.00$_{\pm0.0}$}} & \colorbox[HTML]{67FD9A}{5.67$_{\pm0.6}$} & 19.72$_{\pm0.3}$ &
\colorbox[HTML]{67FD9A}{\underline{79.34$_{\pm0.1}$}} & 25.44$_{\pm0.1}$ & \colorbox[HTML]{67FD9A}{\underline{59.28$_{\pm0.8}$}} & \colorbox[HTML]{67FD9A}{\underline{78.55$_{\pm0.0}$}} & 52.47$_{\pm0.1}$ \\
\hdashline
LoRA & & 3.73/10.71 &
11.00$_{\pm0.0}$ & \colorbox[HTML]{67FD9A}{\textbf{59.00$_{\pm0.0}$}} & \colorbox[HTML]{67FD9A}{\textbf{14.67$_{\pm0.6}$}} & 18.15$_{\pm0.3}$ &
\colorbox[HTML]{67FD9A}{\textbf{79.73$_{\pm0.0}$}} & 33.87$_{\pm0.1}$ & \colorbox[HTML]{67FD9A}{56.32$_{\pm1.0}$} & \colorbox[HTML]{67FD9A}{\textbf{80.07$_{\pm0.0}$}} & 53.63$_{\pm0.2}$ \\
Up-Scaling & & 3.67/10.65 &
13.00$_{\pm0.0}$ & \colorbox[HTML]{67FD9A}{56.00$_{\pm0.0}$} & \colorbox[HTML]{67FD9A}{\underline{11.00$_{\pm1.0}$}} & \textbf{24.78$_{\pm0.3}$} &
\colorbox[HTML]{67FD9A}{78.97$_{\pm0.1}$} & \textbf{47.00$_{\pm1.9}$} & \colorbox[HTML]{67FD9A}{\textbf{59.61$_{\pm0.0}$}} & \colorbox[HTML]{67FD9A}{77.55$_{\pm0.0}$} & \colorbox[HTML]{67FD9A}{\textbf{57.58$_{\pm0.4}$}} \\
Upcycling & \multirow{-4}{*}{Induct} & 3.87/10.85 &
\underline{13.67$_{\pm1.5}$} & \colorbox[HTML]{67FD9A}{53.00$_{\pm0.0}$} & \colorbox[HTML]{67FD9A}{8.67$_{\pm2.1}$} & \underline{24.08$_{\pm0.0}$} &
\colorbox[HTML]{67FD9A}{78.50$_{\pm0.1}$} & \underline{43.61$_{\pm0.3}$} & \colorbox[HTML]{67FD9A}{55.83$_{\pm0.6}$} & \colorbox[HTML]{67FD9A}{76.82$_{\pm0.0}$} & \colorbox[HTML]{67FD9A}{\underline{55.77$_{\pm0.2}$}} \\

\midrule
Full FT & & 8.03/8.03 &
\colorbox[HTML]{67FD9A}{24.67$_{\pm0.6}$} & \colorbox[HTML]{67FD9A}{\textbf{39.00$_{\pm0.0}$}} & \colorbox[HTML]{67FD9A}{\textbf{8.67$_{\pm1.2}$}} & \textbf{26.18$_{\pm0.0}$} &
\colorbox[HTML]{67FD9A}{\textbf{77.24$_{\pm0.1}$}} & 45.92$_{\pm0.7}$ & \colorbox[HTML]{67FD9A}{\textbf{59.61$_{\pm0.9}$}} & \colorbox[HTML]{67FD9A}{\textbf{78.60$_{\pm0.0}$}} & \colorbox[HTML]{67FD9A}{57.51$_{\pm0.1}$} \\
\hdashline
LoRA & & 3.73/10.71 &
\colorbox[HTML]{67FD9A}{31.33$_{\pm1.5}$} & \colorbox[HTML]{67FD9A}{\underline{37.67$_{\pm0.6}$}} & \colorbox[HTML]{67FD9A}{4.00$_{\pm1.0}$} & 24.05$_{\pm0.0}$ &
\colorbox[HTML]{67FD9A}{\underline{77.18$_{\pm0.0}$}} & \colorbox[HTML]{67FD9A}{\underline{52.27$_{\pm0.9}$}} & \colorbox[HTML]{67FD9A}{57.14$_{\pm0.5}$} & \colorbox[HTML]{67FD9A}{\underline{78.38$_{\pm0.0}$}} & \colorbox[HTML]{67FD9A}{\underline{57.81$_{\pm0.3}$}} \\
Up-Scaling & & 3.67/10.65 &
\colorbox[HTML]{67FD9A}{\textbf{37.67$_{\pm1.5}$}} & \colorbox[HTML]{67FD9A}{35.00$_{\pm0.0}$} & \colorbox[HTML]{67FD9A}{4.00$_{\pm0.0}$} & 23.89$_{\pm0.6}$ &
\colorbox[HTML]{67FD9A}{76.34$_{\pm0.0}$} & \colorbox[HTML]{67FD9A}{\textbf{58.92$_{\pm1.1}$}} & \colorbox[HTML]{67FD9A}{\underline{59.28$_{\pm0.6}$}} & \colorbox[HTML]{67FD9A}{75.05$_{\pm0.0}$} & \colorbox[HTML]{67FD9A}{\textbf{58.70$_{\pm0.3}$}} \\
Upcycling & \multirow{-4}{*}{Deduct} & 3.87/10.85 &
\colorbox[HTML]{67FD9A}{\underline{34.00$_{\pm2.0}$}} & \colorbox[HTML]{67FD9A}{33.00$_{\pm0.0}$} & \colorbox[HTML]{67FD9A}{\underline{6.67$_{\pm1.5}$}} & \underline{25.13$_{\pm0.0}$} &
69.80$_{\pm0.4}$ & 20.92$_{\pm0.1}$ & \colorbox[HTML]{67FD9A}{55.34$_{\pm0.6}$} & \colorbox[HTML]{67FD9A}{73.36$_{\pm0.1}$} & 48.91$_{\pm0.0}$ \\

\midrule
Full FT & & 8.03/8.03 &
\underline{13.67$_{\pm1.2}$} & 28.00$_{\pm0.0}$ & \colorbox[HTML]{67FD9A}{5.67$_{\pm1.5}$} & \textbf{21.47$_{\pm0.0}$} &
\colorbox[HTML]{67FD9A}{75.74$_{\pm0.0}$} & \textbf{47.06$_{\pm0.4}$} & \colorbox[HTML]{67FD9A}{\textbf{57.97$_{\pm0.3}$}} & \colorbox[HTML]{67FD9A}{\textbf{81.73$_{\pm0.0}$}} & \colorbox[HTML]{67FD9A}{\textbf{56.79$_{\pm0.1}$}} \\
\hdashline
LoRA & & 3.73/10.71 &
10.33$_{\pm1.5}$ & \colorbox[HTML]{67FD9A}{\underline{35.33$_{\pm0.6}$}} & \colorbox[HTML]{67FD9A}{\textbf{9.00$_{\pm1.0}$}} & 18.85$_{\pm0.0}$ &
\colorbox[HTML]{67FD9A}{75.92$_{\pm0.0}$} & \underline{46.60$_{\pm0.7}$} & \colorbox[HTML]{67FD9A}{\underline{56.65$_{\pm0.0}$}} & \colorbox[HTML]{67FD9A}{\underline{79.87$_{\pm0.0}$}} & \colorbox[HTML]{67FD9A}{55.58$_{\pm0.1}$} \\
Up-Scaling & & 3.67/10.65 &
13.00$_{\pm0.0}$ & \colorbox[HTML]{67FD9A}{\textbf{36.00$_{\pm0.0}$}} & \colorbox[HTML]{67FD9A}{5.67$_{\pm1.5}$} & \underline{21.12$_{\pm0.3}$} &
\colorbox[HTML]{67FD9A}{\underline{77.16$_{\pm0.0}$}} & 46.42$_{\pm0.7}$ & \colorbox[HTML]{67FD9A}{55.34$_{\pm1.2}$} & \colorbox[HTML]{67FD9A}{79.13$_{\pm0.1}$} & \colorbox[HTML]{67FD9A}{\underline{55.83$_{\pm0.4}$}} \\
Upcycling & \multirow{-4}{*}{Abduct} & 3.87/10.85 &
\textbf{14.00$_{\pm0.0}$} & 29.00$_{\pm0.0}$ & \colorbox[HTML]{67FD9A}{\underline{8.00$_{\pm1.0}$}} & 19.90$_{\pm0.0}$ &
\colorbox[HTML]{67FD9A}{\textbf{77.37$_{\pm0.0}$}} & 32.02$_{\pm0.1}$ & \colorbox[HTML]{67FD9A}{56.16$_{\pm1.7}$} & \colorbox[HTML]{67FD9A}{78.16$_{\pm0.1}$} & 52.72$_{\pm0.4}$ \\

\midrule
Full FT &  & 8.03/8.03 &
\colorbox[HTML]{67FD9A}{26.00$_{\pm0.0}$} & 0.00$_{\pm0.0}$ & \colorbox[HTML]{67FD9A}{5.33$_{\pm0.6}$} &
18.85$_{\pm0.0}$ & \colorbox[HTML]{67FD9A}{76.03$_{\pm0.1}$} & 29.93$_{\pm0.9}$ &
\colorbox[HTML]{67FD9A}{55.83$_{\pm0.3}$} & \colorbox[HTML]{67FD9A}{\textbf{80.51$_{\pm0.0}$}} & 52.23$_{\pm0.2}$ \\
\hdashline
LoRA &  & 3.73/10.71 &
\colorbox[HTML]{67FD9A}{28.33$_{\pm3.2}$} & 0.00$_{\pm0.0}$ & \colorbox[HTML]{67FD9A}{\underline{8.33$_{\pm0.6}$}} &
\underline{21.12$_{\pm0.6}$} & \colorbox[HTML]{67FD9A}{\underline{78.22$_{\pm0.2}$}} & \underline{43.78$_{\pm4.2}$} &
\colorbox[HTML]{67FD9A}{\underline{56.32$_{\pm1.6}$}} & \colorbox[HTML]{67FD9A}{\underline{79.76$_{\pm1.1}$}} & \colorbox[HTML]{67FD9A}{\underline{55.84$_{\pm0.8}$}} \\
Up-Scaling &  & 3.67/10.65 &
\colorbox[HTML]{67FD9A}{\underline{30.00$_{\pm3.5}$}} & 0.00$_{\pm0.0}$ & \colorbox[HTML]{67FD9A}{\textbf{9.67$_{\pm1.5}$}} &
20.60$_{\pm1.2}$ & \colorbox[HTML]{67FD9A}{\textbf{78.25$_{\pm0.2}$}} & \textbf{45.58$_{\pm4.6}$} &
\colorbox[HTML]{67FD9A}{\textbf{56.82$_{\pm1.4}$}} & \colorbox[HTML]{67FD9A}{79.09$_{\pm1.1}$} & \colorbox[HTML]{67FD9A}{\textbf{56.07$_{\pm0.8}$}} \\
Upcycling & \multirow{-4}{*}{Mix} & 3.87/10.85 &
\colorbox[HTML]{67FD9A}{\textbf{32.33$_{\pm0.6}$}} & \colorbox[HTML]{67FD9A}{\textbf{56.00$_{\pm0.0}$}} & \colorbox[HTML]{67FD9A}{4.67$_{\pm2.1}$} &
\textbf{26.18$_{\pm0.0}$} & \colorbox[HTML]{67FD9A}{77.02$_{\pm0.0}$} & 43.63$_{\pm0.6}$ &
\colorbox[HTML]{67FD9A}{54.19$_{\pm0.0}$} & \colorbox[HTML]{67FD9A}{77.33$_{\pm0.0}$} & \colorbox[HTML]{67FD9A}{55.67$_{\pm0.1}$} \\

\midrule\midrule
\rowcolor{gray!20} Llama-3.3-70B-IT & - & 0/70.0 &
33.00$_{\pm0.3}$ & 55.33$_{\pm2.5}$ & 16.00$_{\pm2.6}$ &
39.96$_{\pm0.3}$ & 79.28$_{\pm0.2}$ & 65.75$_{\pm0.2}$ &
60.59$_{\pm0.9}$ & 77.07$_{\pm0.1}$ & 64.53$_{\pm0.2}$ \\
\rowcolor{gray!20} Qwen3-8B & - & 0/8.19 &
79.67$_{\pm2.1}$ & 67.33$_{\pm0.6}$ & 35.33$_{\pm0.6}$ & 40.14$_{\pm0.8}$ &
81.26$_{\pm0.0}$ & 70.98$_{\pm0.7}$ & 78.33$_{\pm1.0}$ & 71.40$_{\pm0.0}$ & 68.42$_{\pm0.2}$ \\
\midrule
Full FT & & 8.19/8.19 &
\underline{68.67$_{\pm0.6}$} & 58.00$_{\pm0.0}$ & 24.67$_{\pm1.2}$ & 32.98$_{\pm2.7}$ &
78.90$_{\pm0.0}$ & 55.68$_{\pm1.4}$ & \underline{75.21$_{\pm0.3}$} & \colorbox[HTML]{67FD9A}{71.91$_{\pm0.0}$} & 62.94$_{\pm0.9}$ \\
\hdashline
LoRA & & 3.88/10.83 &
61.67$_{\pm0.6}$ & \underline{60.00$_{\pm0.0}$} & 14.33$_{\pm0.6}$ & 30.89$_{\pm0.9}$ &
78.25$_{\pm0.0}$ & 39.07$_{\pm1.6}$ & 67.16$_{\pm0.3}$ & \colorbox[HTML]{67FD9A}{72.18$_{\pm0.0}$} & 57.51$_{\pm0.5}$ \\
Up-Scaling & & 3.56/10.51 &
34.00$_{\pm1.7}$ & 57.00$_{\pm0.0}$ & \colorbox[HTML]{67FD9A}{\textbf{40.67$_{\pm0.6}$}} & \underline{33.86$_{\pm5.1}$} &
\colorbox[HTML]{67FD9A}{\textbf{82.06$_{\pm0.0}$}} & \underline{67.60$_{\pm1.0}$} & 73.89$_{\pm0.5}$ & \colorbox[HTML]{67FD9A}{\textbf{72.67$_{\pm0.0}$}} & \underline{66.02$_{\pm1.1}$} \\
Upcycling & \multirow{-4}{*}{Induct} & 3.96/10.91 &
\textbf{72.67$_{\pm1.5}$} & \colorbox[HTML]{67FD9A}{\textbf{72.00$_{\pm0.0}$}} & \underline{31.67$_{\pm1.2}$} & \colorbox[HTML]{67FD9A}{\textbf{41.19$_{\pm2.4}$}} &
\colorbox[HTML]{67FD9A}{\underline{81.88$_{\pm0.0}$}} & \colorbox[HTML]{67FD9A}{\textbf{74.54$_{\pm1.0}$}} & \textbf{77.67$_{\pm0.3}$} & \colorbox[HTML]{67FD9A}{\underline{72.20$_{\pm0.0}$}} & \colorbox[HTML]{67FD9A}{\textbf{69.49$_{\pm0.7}$}} \\

\midrule
Full FT & & 8.19/8.19 &
\underline{35.33$_{\pm0.6}$} & \underline{64.33$_{\pm0.6}$} & \textbf{19.33$_{\pm0.6}$} & 34.21$_{\pm3.3}$ &
\colorbox[HTML]{67FD9A}{81.55$_{\pm0.0}$} & \underline{67.26$_{\pm0.4}$} & 71.76$_{\pm0.3}$ & 70.53$_{\pm0.0}$ & \underline{65.06$_{\pm0.6}$} \\
\hdashline
LoRA & & 3.88/10.83 &
28.00$_{\pm1.0}$ & 58.33$_{\pm1.2}$ & \underline{15.00$_{\pm2.6}$} & 31.76$_{\pm1.2}$ &
\colorbox[HTML]{67FD9A}{\textbf{83.25$_{\pm0.0}$}} & 55.26$_{\pm1.1}$ & \colorbox[HTML]{67FD9A}{\underline{80.30$_{\pm0.0}$}} & \underline{70.96$_{\pm0.0}$} & 64.31$_{\pm0.0}$ \\
Up-Scaling & & 3.56/10.51 &
28.33$_{\pm0.6}$ & 62.00$_{\pm0.0}$ & 12.00$_{\pm0.0}$ & \underline{36.82$_{\pm5.1}$} &
74.29$_{\pm0.1}$ & 52.60$_{\pm1.0}$ & \colorbox[HTML]{67FD9A}{\textbf{82.43$_{\pm0.3}$}} & 70.00$_{\pm0.0}$ & 63.23$_{\pm0.8}$ \\
Upcycling & \multirow{-4}{*}{Deduct} & 3.96/10.91 &
\textbf{72.00$_{\pm0.0}$} & \colorbox[HTML]{67FD9A}{\textbf{71.00$_{\pm0.0}$}} & 12.00$_{\pm1.0}$ & \colorbox[HTML]{67FD9A}{\textbf{42.76$_{\pm2.1}$}} &
\colorbox[HTML]{67FD9A}{\underline{82.77$_{\pm0.0}$}} & \colorbox[HTML]{67FD9A}{\textbf{73.59$_{\pm0.9}$}} & \colorbox[HTML]{67FD9A}{80.13$_{\pm0.3}$} & \colorbox[HTML]{67FD9A}{\textbf{71.45$_{\pm0.0}$}} & \colorbox[HTML]{67FD9A}{\textbf{70.14$_{\pm0.4}$}} \\

\midrule
Full FT & & 8.19/8.19 &
33.33$_{\pm1.2}$ & \colorbox[HTML]{67FD9A}{\textbf{69.33$_{\pm0.6}$}} & \textbf{17.33$_{\pm1.2}$} & \colorbox[HTML]{67FD9A}{\underline{40.31$_{\pm0.0}$}} &
\colorbox[HTML]{67FD9A}{\textbf{83.81$_{\pm0.1}$}} & \underline{67.79$_{\pm0.7}$} & 71.10$_{\pm0.3}$ & \colorbox[HTML]{67FD9A}{\underline{73.96$_{\pm0.0}$}} & \underline{67.39$_{\pm0.1}$} \\
\hdashline
LoRA & & 3.88/10.83 &
43.67$_{\pm0.6}$ & 64.00$_{\pm0.0}$ & \underline{16.00$_{\pm2.0}$} & 39.61$_{\pm0.8}$ &
\colorbox[HTML]{67FD9A}{81.96$_{\pm0.1}$} & 53.33$_{\pm1.2}$ & 71.76$_{\pm0.7}$ & \colorbox[HTML]{67FD9A}{\textbf{74.11$_{\pm0.0}$}} & 64.16$_{\pm0.2}$ \\
Up-Scaling & & 3.56/10.51 &
\underline{44.33$_{\pm0.6}$} & 60.00$_{\pm0.0}$ & 14.33$_{\pm1.2}$ & 37.00$_{\pm0.3}$ &
\colorbox[HTML]{67FD9A}{81.92$_{\pm0.0}$} & 44.09$_{\pm0.9}$ & \underline{72.25$_{\pm0.3}$} & \colorbox[HTML]{67FD9A}{73.47$_{\pm0.0}$} & 61.74$_{\pm0.1}$ \\
Upcycling & \multirow{-4}{*}{Abduct} & 3.96/10.91 &
\textbf{53.33$_{\pm0.6}$} & \colorbox[HTML]{67FD9A}{\underline{68.33$_{\pm0.6}$}} & 13.67$_{\pm0.6}$ & \colorbox[HTML]{67FD9A}{\textbf{47.82$_{\pm0.6}$}} &
\colorbox[HTML]{67FD9A}{\underline{83.74$_{\pm0.0}$}} & \textbf{68.08$_{\pm0.9}$} & \textbf{75.53$_{\pm0.3}$} & \colorbox[HTML]{67FD9A}{71.51$_{\pm0.1}$} & \colorbox[HTML]{67FD9A}{\textbf{69.34$_{\pm0.1}$}} \\
\midrule
Full FT &  & 8.19/8.19 &
31.00$_{\pm0.0}$ & 60.00$_{\pm0.0}$ & \textbf{20.33$_{\pm0.6}$} &
\underline{35.78$_{\pm0.6}$} & \colorbox[HTML]{67FD9A}{\textbf{82.34$_{\pm0.0}$}} & \underline{62.84}$_{\pm0.1}$ &
\underline{73.56$_{\pm0.3}$} & \colorbox[HTML]{67FD9A}{72.40$_{\pm0.0}$} & \underline{65.38$_{\pm0.2}$} \\
\hdashline
LoRA &  &  3.88/10.83 &
\underline{33.00$_{\pm1.0}$} & 61.33$_{\pm2.3}$ & 11.67$_{\pm2.5}$ &
31.94$_{\pm0.0}$ & \colorbox[HTML]{67FD9A}{81.52$_{\pm0.2}$} & 59.56$_{\pm7.5}$ &
66.50$_{\pm5.6}$ & \colorbox[HTML]{67FD9A}{\underline{73.82$_{\pm0.3}$}} & 62.67$_{\pm0.4}$ \\
Up-Scaling &  & 3.56/10.51 &
\underline{33.00$_{\pm1.7}$} & \underline{62.33$_{\pm2.1}$} & \underline{12.67$_{\pm2.1}$} &
31.94$_{\pm0.0}$ & \colorbox[HTML]{67FD9A}{81.42$_{\pm0.2}$} & 53.63$_{\pm8.8}$ &
69.29$_{\pm5.8}$ & \colorbox[HTML]{67FD9A}{\textbf{73.98$_{\pm0.3}$}} & 62.05$_{\pm0.6}$ \\
Upcycling & \multirow{-4}{*}{Mix} & 3.96/10.91 &
\textbf{48.33$_{\pm0.6}$} & \textbf{64.33$_{\pm0.6}$} & 12.00$_{\pm1.0}$ &
\colorbox[HTML]{67FD9A}{\textbf{43.45$_{\pm0.9}$}} & \colorbox[HTML]{67FD9A}{\underline{81.89$_{\pm0.1}$}} & \colorbox[HTML]{67FD9A}{\textbf{71.13$_{\pm0.7}$}} &
\textbf{77.34$_{\pm0.0}$} & 71.40$_{\pm0.0}$ & \colorbox[HTML]{67FD9A}{\textbf{69.04$_{\pm0.3}$}} \\
\bottomrule
\end{tabular}
\end{adjustbox}
\end{center}
\caption{Accuracy of the fine-tuned models on the fundamental reasoning tasks using trajectories collected from \textit{Llama-3.3-70B-Instruct}. \textbf{Method} indicates the base model and the applied PEFT methods. \textbf{FT} shows the fundamental reasoning task used for fine-tuning. \textbf{Train/Total} denotes the number of trainable and total parameters. The best and second-best methods for each fundamental reasoning task are indicated in \textbf{bold} and \underline{underlined}, respectively. Scores that exceed the \textit{base} model performance are shown in \colorbox[HTML]{67FD9A}{green}.}
\label{tab:results}
\end{table*}

\begin{table*}[ht]
\begin{center}
\small
\setlength{\tabcolsep}{1mm}
\begin{adjustbox}{width=0.9\textwidth}
\begin{tabular}{lccccccccccc}
\toprule
& & & \multicolumn{3}{c}{\textbf{Symbolic Tasks}} & \multicolumn{6}{c}{\textbf{Realistic Tasks}} \\
\cmidrule(l{2pt}r{2pt}){4-6} \cmidrule(l{2pt}r{2pt}){7-12}
\textbf{Method} & \textbf{FT} & \textbf{Train/Total} &
\textbf{Deduct} & \textbf{Induct} & \textbf{Abduct} &
\textbf{Detect} & \textbf{$\alpha$NLI} & \textbf{Winowhy} &
\textbf{Folio} & \textbf{RECV} & \textbf{Avg} \\
\midrule
\rowcolor{gray!20} Qwen3-30B & - & 0/30.5 &
82.00$_{\pm1.0}$ & 81.00$_{\pm4.0}$ & 74.33$_{\pm4.2}$ &
33.51$_{\pm3.2}$ & 75.77$_{\pm0.1}$ & 72.73$_{\pm0.2}$ &
72.41$_{\pm0.5}$ & 80.56$_{\pm0.3}$ & 67.00$_{\pm0.6}$ \\
\rowcolor{gray!20} Llama-3.1-8B-IT & - & 0/8.03 &
14.00$_{\pm1.0}$ & 32.00$_{\pm0.0}$ & 1.67$_{\pm0.6}$ & 28.80$_{\pm0.0}$ &
73.97$_{\pm0.0}$ & 48.60$_{\pm0.4}$ & 53.69$_{\pm0.5}$ & 71.98$_{\pm0.0}$ & 55.41$_{\pm0.0}$ \\
\midrule
Full FT &  & 8.03/8.03 &
\colorbox[HTML]{67FD9A}{\underline{25.67$_{\pm1.5}$}} & \colorbox[HTML]{67FD9A}{\textbf{78.00$_{\pm0.0}$}} & \colorbox[HTML]{67FD9A}{2.33$_{\pm0.6}$} & \textbf{25.13$_{\pm0.0}$} &
\colorbox[HTML]{67FD9A}{77.71$_{\pm0.0}$} & 43.05$_{\pm0.1}$ & \colorbox[HTML]{67FD9A}{54.84$_{\pm0.3}$} & \colorbox[HTML]{67FD9A}{\textbf{78.29$_{\pm0.0}$}} & \colorbox[HTML]{67FD9A}{\underline{55.80$_{\pm0.1}$}} \\
\hdashline
LoRA &  & 3.73/10.71 &
\colorbox[HTML]{67FD9A}{24.33$_{\pm1.2}$} & \colorbox[HTML]{67FD9A}{\underline{69.33$_{\pm0.6}$}} & 1.67$_{\pm1.2}$ & 21.82$_{\pm0.3}$ &
\colorbox[HTML]{67FD9A}{\textbf{78.23$_{\pm0.0}$}} & \underline{44.33$_{\pm0.4}$} & \colorbox[HTML]{67FD9A}{56.00$_{\pm0.3}$} & \colorbox[HTML]{67FD9A}{\underline{76.82$_{\pm0.0}$}} & \colorbox[HTML]{67FD9A}{55.44$_{\pm0.1}$} \\
Up-Scaling &  & 3.67/10.65 &
\colorbox[HTML]{67FD9A}{34.67$_{\pm1.5}$} & \colorbox[HTML]{67FD9A}{\underline{69.33$_{\pm0.6}$}} & \colorbox[HTML]{67FD9A}{\underline{5.33$_{\pm1.2}$}} & \underline{24.43$_{\pm0.3}$} &
67.37$_{\pm0.1}$ & 35.25$_{\pm0.1}$ & \colorbox[HTML]{67FD9A}{\textbf{56.98$_{\pm0.6}$}} & \colorbox[HTML]{67FD9A}{72.36$_{\pm0.5}$} & 51.28$_{\pm0.3}$ \\
Upcycling & \multirow{-4}{*}{Induct} & 3.87/10.85 &
\colorbox[HTML]{67FD9A}{\textbf{26.00$_{\pm1.0}$}} & \colorbox[HTML]{67FD9A}{64.00$_{\pm0.0}$} & \colorbox[HTML]{67FD9A}{\textbf{5.67$_{\pm1.5}$}} & \colorbox[HTML]{67FD9A}{24.08$_{\pm0.0}$} &
\colorbox[HTML]{67FD9A}{\underline{78.18$_{\pm0.1}$}} & \colorbox[HTML]{67FD9A}{\textbf{51.67$_{\pm0.3}$}} & 51.96$_{\pm2.6}$ & \colorbox[HTML]{67FD9A}{74.73$_{\pm0.0}$} & \colorbox[HTML]{67FD9A}{\textbf{56.13$_{\pm0.5}$}} \\

\midrule
Full FT &  & 8.03/8.03 &
\colorbox[HTML]{67FD9A}{\underline{67.33$_{\pm0.6}$}} & \colorbox[HTML]{67FD9A}{33.00$_{\pm0.0}$} & \colorbox[HTML]{67FD9A}{\textbf{7.67$_{\pm1.5}$}} &
24.61$_{\pm0.0}$ & 48.73$_{\pm0.1}$ & \underline{41.51$_{\pm0.3}$} &
\colorbox[HTML]{67FD9A}{\underline{56.65$_{\pm0.0}$}} & \colorbox[HTML]{67FD9A}{\textbf{77.80$_{\pm0.0}$}} & 49.86$_{\pm0.1}$ \\
\hdashline
LoRA &  & 3.73/10.71 &
\colorbox[HTML]{67FD9A}{39.00$_{\pm3.5}$} & \colorbox[HTML]{67FD9A}{\underline{36.00$_{\pm0.0}$}} & \colorbox[HTML]{67FD9A}{6.00$_{\pm1.0}$} &
\colorbox[HTML]{67FD9A}{\textbf{29.49$_{\pm0.3}$}} & \textbf{70.35$_{\pm0.2}$} & 40.42$_{\pm0.4}$ &
\colorbox[HTML]{67FD9A}{\underline{56.65$_{\pm0.0}$}} & \colorbox[HTML]{67FD9A}{\underline{77.02$_{\pm0.0}$}} & \textbf{54.79$_{\pm0.1}$} \\
Up-Scaling &  & 3.67/10.65 &
\colorbox[HTML]{67FD9A}{\textbf{70.00$_{\pm1.7}$}} & 32.00$_{\pm0.0}$ & \colorbox[HTML]{67FD9A}{3.33$_{\pm1.5}$} &
\underline{27.23$_{\pm0.0}$} & 57.26$_{\pm0.2}$ & \textbf{42.93$_{\pm0.3}$} &
\colorbox[HTML]{67FD9A}{\textbf{58.46$_{\pm0.3}$}} & \colorbox[HTML]{67FD9A}{73.87$_{\pm0.0}$} & 51.95$_{\pm0.1}$ \\
Upcycling & \multirow{-4}{*}{Deduct}  & 3.87/10.85 &
\colorbox[HTML]{67FD9A}{47.00$_{\pm1.0}$} & \colorbox[HTML]{67FD9A}{\textbf{40.00$_{\pm0.0}$}} & \colorbox[HTML]{67FD9A}{\underline{6.33$_{\pm1.5}$}} &
26.18$_{\pm0.0}$ & \underline{69.87$_{\pm0.1}$} & 37.69$_{\pm0.1}$ &
\colorbox[HTML]{67FD9A}{\underline{56.65$_{\pm0.0}$}} & \colorbox[HTML]{67FD9A}{73.29$_{\pm0.0}$} & \underline{52.74$_{\pm0.0}$} \\

\midrule
Full FT &  & 8.03/8.03 &
\colorbox[HTML]{67FD9A}{\underline{34.67$_{\pm1.5}$}} & 28.00$_{\pm0.0}$ & \colorbox[HTML]{67FD9A}{\underline{39.67$_{\pm0.6}$}} &
18.67$_{\pm0.3}$ & \colorbox[HTML]{67FD9A}{\textbf{77.27$_{\pm0.1}$}} & \colorbox[HTML]{67FD9A}{\textbf{63.20$_{\pm0.8}$}} &
\colorbox[HTML]{67FD9A}{59.61$_{\pm0.5}$} & \colorbox[HTML]{67FD9A}{\textbf{77.20$_{\pm0.1}$}} & \colorbox[HTML]{67FD9A}{\textbf{59.19$_{\pm0.3}$}} \\
\hdashline
LoRA &  & 3.73/10.71 &
\colorbox[HTML]{67FD9A}{\underline{34.67$_{\pm2.9}$}} & 29.33$_{\pm1.2}$ & \colorbox[HTML]{67FD9A}{\textbf{43.33$_{\pm1.5}$}} &
\underline{23.56$_{\pm0.0}$} & \colorbox[HTML]{67FD9A}{\underline{75.59$_{\pm0.1}$}} & \colorbox[HTML]{67FD9A}{\underline{56.74$_{\pm2.9}$}} &
\colorbox[HTML]{67FD9A}{\underline{59.77$_{\pm0.8}$}} &\colorbox[HTML]{67FD9A}{ 75.09$_{\pm0.0}$} & \colorbox[HTML]{67FD9A}{\underline{58.15$_{\pm0.7}$}} \\
Up-Scaling
&  & 3.67/10.65 &
\colorbox[HTML]{67FD9A}{34.33$_{\pm0.6}$} & \colorbox[HTML]{67FD9A}{\textbf{39.00$_{\pm0.0}$}} & \colorbox[HTML]{67FD9A}{39.33$_{\pm1.2}$} &
23.04$_{\pm0.0}$ & 70.57$_{\pm0.0}$ & 48.19$_{\pm0.2}$ &
52.38$_{\pm0.3}$ & 70.89$_{\pm0.0}$ & 53.01$_{\pm0.0}$ \\

Upcycling & \multirow{-4}{*}{Abduct}  & 3.87/10.85 &
\colorbox[HTML]{67FD9A}{\textbf{41.67$_{\pm0.6}$}} & \colorbox[HTML]{67FD9A}{\underline{36.00$_{\pm0.0}$}} & \colorbox[HTML]{67FD9A}{35.00$_{\pm2.6}$} &
\textbf{26.70$_{\pm0.0}$} & 68.02$_{\pm0.1}$ & \colorbox[HTML]{67FD9A}{51.76$_{\pm0.3}$} &
\colorbox[HTML]{67FD9A}{\textbf{63.05$_{\pm0.5}$}} & \colorbox[HTML]{67FD9A}{\underline{75.73$_{\pm0.1}$}} & \colorbox[HTML]{67FD9A}{57.05$_{\pm0.1}$} \\

\midrule
Full FT &  & 8.03/8.03 &
\colorbox[HTML]{67FD9A}{50.67$_{\pm0.6}$} & \colorbox[HTML]{67FD9A}{63.33$_{\pm0.6}$} & \colorbox[HTML]{67FD9A}{\textbf{45.00$_{\pm2.0}$}} &
\textbf{24.08$_{\pm0.0}$} & \colorbox[HTML]{67FD9A}{\underline{77.54$_{\pm0.0}$}} & \colorbox[HTML]{67FD9A}{\textbf{55.17$_{\pm0.1}$}} &
\colorbox[HTML]{67FD9A}{57.64$_{\pm0.5}$} & \colorbox[HTML]{67FD9A}{\textbf{78.24$_{\pm0.0}$}} & \colorbox[HTML]{67FD9A}{\textbf{58.53$_{\pm0.1}$}} \\
\hdashline
LoRA &  & 3.73/10.71 &
\colorbox[HTML]{67FD9A}{\textbf{69.33$_{\pm2.3}$}} & \colorbox[HTML]{67FD9A}{\textbf{65.67$_{\pm1.2}$}} & \colorbox[HTML]{67FD9A}{\underline{43.67$_{\pm1.2}$}} &
23.39$_{\pm0.3}$ & \colorbox[HTML]{67FD9A}{\textbf{79.26$_{\pm0.1}$}} & 44.89$_{\pm0.2}$ &
\colorbox[HTML]{67FD9A}{\textbf{58.29$_{\pm0.7}$}} & \colorbox[HTML]{67FD9A}{\underline{77.67$_{\pm0.0}$}} & \colorbox[HTML]{67FD9A}{\underline{56.70$_{\pm0.2}$}} \\
Up-Scaling &  & 3.67/10.65 &
\colorbox[HTML]{67FD9A}{\underline{62.00$_{\pm1.0}$}} & \colorbox[HTML]{67FD9A}{\underline{65.00$_{\pm0.0}$}} & \colorbox[HTML]{67FD9A}{38.33$_{\pm1.2}$} &
\colorbox[HTML]{67FD9A}{\textbf{24.08$_{\pm0.0}$}} & \colorbox[HTML]{67FD9A}{76.57$_{\pm0.0}$} & \colorbox[HTML]{67FD9A}{49.59$_{\pm0.2}$} &
\colorbox[HTML]{67FD9A}{56.49$_{\pm0.7}$} & \colorbox[HTML]{67FD9A}{75.82$_{\pm0.0}$} & \colorbox[HTML]{67FD9A}{56.51$_{\pm0.2}$} \\
Upcycling & \multirow{-4}{*}{Mix}  & 3.87/10.85 &
\colorbox[HTML]{67FD9A}{58.00$_{\pm0.0}$} & \colorbox[HTML]{67FD9A}{58.33$_{\pm0.6}$} & \colorbox[HTML]{67FD9A}{34.33$_{\pm2.9}$} &
21.99$_{\pm0.0}$ & \colorbox[HTML]{67FD9A}{74.36$_{\pm0.1}$} & \colorbox[HTML]{67FD9A}{\underline{52.23$_{\pm0.1}$}} &
\colorbox[HTML]{67FD9A}{\underline{58.13$_{\pm0.5}$}} & \colorbox[HTML]{67FD9A}{74.16$_{\pm0.1}$} & \colorbox[HTML]{67FD9A}{56.17$_{\pm0.1}$} \\

\midrule\midrule
\rowcolor{gray!20} Qwen3-30B & - & 0/30.5 &
82.00$_{\pm1.0}$ & 81.00$_{\pm4.0}$ & 74.33$_{\pm4.2}$ &
33.51$_{\pm3.2}$ & 75.77$_{\pm0.1}$ & 72.73$_{\pm0.2}$ &
72.41$_{\pm0.5}$ & 80.56$_{\pm0.3}$ & 67.00$_{\pm0.6}$ \\
\rowcolor{gray!20} Qwen3-8B & - & 0/8.19 &
79.67$_{\pm2.1}$ & 67.33$_{\pm0.6}$ & 35.33$_{\pm0.6}$ & 40.14$_{\pm0.8}$ &
81.26$_{\pm0.0}$ & 70.98$_{\pm0.7}$ & 78.33$_{\pm1.0}$ & 71.40$_{\pm0.0}$ & 68.42$_{\pm0.2}$ \\

\midrule
Full FT &  & 8.19/8.19 &
68.33$_{\pm0.6}$ & 49.00$_{\pm0.0}$ & \underline{7.67$_{\pm0.6}$} & 27.23$_{\pm0.0}$ &
\colorbox[HTML]{67FD9A}{\underline{84.57$_{\pm0.0}$}} & \colorbox[HTML]{67FD9A}{\textbf{79.34$_{\pm0.2}$}} & 77.18$_{\pm0.3}$ & 70.42$_{\pm0.0}$ & 67.75$_{\pm0.1}$ \\
\hdashline
LoRA &  & 3.88/10.83 &
\colorbox[HTML]{67FD9A}{\underline{82.67$_{\pm0.6}$}} & \colorbox[HTML]{67FD9A}{\underline{75.33$_{\pm1.2}$}} & 2.67$_{\pm0.6}$ & \colorbox[HTML]{67FD9A}{\underline{40.84$_{\pm0.9}$}} &
\colorbox[HTML]{67FD9A}{\textbf{85.56$_{\pm0.1}$}} & \colorbox[HTML]{67FD9A}{\underline{77.88$_{\pm0.1}$}} & 77.83$_{\pm0.5}$ & \underline{70.64$_{\pm0.0}$} & \colorbox[HTML]{67FD9A}{\textbf{70.55$_{\pm0.2}$}} \\
Up-Scaling &  & 3.56/10.51 &
6.33$_{\pm0.6}$ & 41.00$_{\pm0.0}$ & 0.00$_{\pm0.0}$ & \colorbox[HTML]{67FD9A}{\textbf{43.63$_{\pm0.6}$}} &
\colorbox[HTML]{67FD9A}{83.53$_{\pm0.1}$} & 56.00$_{\pm0.6}$ & \colorbox[HTML]{67FD9A}{\underline{78.49$_{\pm0.3}$}} & 70.18$_{\pm0.2}$ & 66.37$_{\pm0.0}$ \\
Upcycling & \multirow{-4}{*}{Induct} & 3.96/10.91 &
\colorbox[HTML]{67FD9A}{\textbf{89.00$_{\pm0.0}$}} & \colorbox[HTML]{67FD9A}{\textbf{77.00$_{\pm1.0}$}} & \textbf{11.00$_{\pm1.0}$} & 40.14$_{\pm1.1}$ &
80.34$_{\pm0.2}$ & \colorbox[HTML]{67FD9A}{75.65$_{\pm0.2}$} & \colorbox[HTML]{67FD9A}{\textbf{84.24$_{\pm0.5}$}} & \textbf{71.13$_{\pm0.0}$} & \colorbox[HTML]{67FD9A}{\underline{70.30$_{\pm0.3}$}} \\

\midrule
Full FT &  & 8.19/8.19 &
\colorbox[HTML]{67FD9A}{\underline{88.00$_{\pm1.7}$}} & 60.00$_{\pm0.0}$ & \colorbox[HTML]{67FD9A}{\textbf{48.00$_{\pm1.0}$}} &
36.65$_{\pm0.0}$ & \colorbox[HTML]{67FD9A}{81.92$_{\pm0.0}$} & \underline{70.30$_{\pm0.4}$} &
75.04$_{\pm0.8}$ & 70.98$_{\pm0.0}$ & 66.98$_{\pm0.2}$ \\
\hdashline
LoRA &  & 3.88/10.83 &
\colorbox[HTML]{67FD9A}{84.00$_{\pm1.0}$} & \colorbox[HTML]{67FD9A}{\underline{68.67$_{\pm1.2}$}} & 21.33$_{\pm1.5}$ &
\colorbox[HTML]{67FD9A}{\underline{42.76$_{\pm0.3}$}} & \colorbox[HTML]{67FD9A}{\underline{82.70$_{\pm0.1}$}} & 69.83$_{\pm0.2}$ &
\colorbox[HTML]{67FD9A}{\underline{80.79$_{\pm1.0}$}} & \underline{71.13$_{\pm0.0}$} & \colorbox[HTML]{67FD9A}{\underline{69.44$_{\pm0.3}$}} \\
Up-Scaling &  & 3.56/10.51 &
\colorbox[HTML]{67FD9A}{85.33$_{\pm1.2}$} & 63.33$_{\pm0.6}$ & \underline{34.67$_{\pm2.1}$} &
37.35$_{\pm0.6}$ & 80.84$_{\pm0.1}$ & 19.82$_{\pm0.1}$ &
\colorbox[HTML]{67FD9A}{80.30$_{\pm0.0}$} & \textbf{71.40$_{\pm0.0}$} & 57.94$_{\pm0.1}$ \\
Upcycling & \multirow{-4}{*}{Deduct}  & 3.96/10.91 &
\colorbox[HTML]{67FD9A}{\textbf{92.00$_{\pm1.0}$}} & \colorbox[HTML]{67FD9A}{\textbf{73.00$_{\pm0.0}$}} & \underline{34.67$_{\pm0.6}$} &
\colorbox[HTML]{67FD9A}{\textbf{44.50$_{\pm0.5}$}} & \colorbox[HTML]{67FD9A}{\textbf{84.00$_{\pm0.0}$}} & \colorbox[HTML]{67FD9A}{\textbf{73.58$_{\pm0.2}$}} &
\colorbox[HTML]{67FD9A}{\textbf{81.94$_{\pm0.3}$}} & 70.82$_{\pm0.1}$ & \colorbox[HTML]{67FD9A}{\textbf{70.97$_{\pm0.2}$}} \\

\midrule
Full FT &  & 8.19/8.19 &
74.33$_{\pm0.6}$ & \colorbox[HTML]{67FD9A}{\underline{67.67$_{\pm0.6}$}} & 32.00$_{\pm2.0}$ &
\textbf{38.57$_{\pm0.3}$} & \colorbox[HTML]{67FD9A}{81.54$_{\pm0.1}$} & \textbf{69.03$_{\pm0.8}$} &
\underline{77.18$_{\pm0.3}$} & 69.82$_{\pm0.2}$ & \textbf{67.23$_{\pm0.1}$} \\
\hdashline
LoRA &  & 3.88/10.83 &
\textbf{77.67$_{\pm1.5}$} & \colorbox[HTML]{67FD9A}{\textbf{70.67$_{\pm0.6}$}} & \colorbox[HTML]{67FD9A}{\underline{41.33$_{\pm1.5}$}} &
\underline{36.13$_{\pm0.0}$} & \colorbox[HTML]{67FD9A}{\underline{81.97$_{\pm0.1}$}} & 66.72$_{\pm0.6}$ &
76.03$_{\pm1.0}$ & \underline{70.02$_{\pm0.0}$} & 66.18$_{\pm0.1}$ \\
Up-Scaling &  & 3.56/10.51 &
68.67$_{\pm0.6}$ & 62.33$_{\pm0.6}$ & \colorbox[HTML]{67FD9A}{35.67$_{\pm0.6}$} &
34.20$_{\pm0.3}$ & 76.05$_{\pm0.1}$ & 27.05$_{\pm0.0}$ &
70.44$_{\pm0.0}$ & 68.91$_{\pm0.1}$ & 55.33$_{\pm0.1}$ \\

Upcycling & \multirow{-4}{*}{Abduct} & 3.96/10.91 &
\underline{74.67$_{\pm1.2}$} & 62.00$_{\pm0.0}$ & \colorbox[HTML]{67FD9A}{\textbf{46.00$_{\pm1.7}$}} &
33.86$_{\pm0.3}$ & \colorbox[HTML]{67FD9A}{\textbf{82.73$_{\pm0.0}$}} & \underline{68.20$_{\pm0.6}$} &
\colorbox[HTML]{67FD9A}{\textbf{79.15$_{\pm0.3}$}} & \textbf{70.49$_{\pm0.0}$} & \underline{66.89$_{\pm0.1}$} \\
\midrule
Full FT &  & 8.19/8.19 &
\colorbox[HTML]{67FD9A}{\textbf{88.33$_{\pm1.2}$}} & \colorbox[HTML]{67FD9A}{\textbf{77.33$_{\pm0.6}$}} & \colorbox[HTML]{67FD9A}{\underline{41.33$_{\pm1.2}$}} &
39.44$_{\pm0.3}$ & \colorbox[HTML]{67FD9A}{81.69$_{\pm0.0}$} & \colorbox[HTML]{67FD9A}{\textbf{73.44$_{\pm0.2}$}} &
74.55$_{\pm0.3}$ & \underline{70.29$_{\pm0.0}$} & \underline{67.88$_{\pm0.1}$} \\
\hdashline
LoRA &  & 3.88/10.83 &
\colorbox[HTML]{67FD9A}{85.33$_{\pm1.5}$} & \colorbox[HTML]{67FD9A}{74.67$_{\pm0.6}$} & 23.67$_{\pm3.1}$ &
\colorbox[HTML]{67FD9A}{40.32$_{\pm0.9}$} & \colorbox[HTML]{67FD9A}{\underline{84.10$_{\pm0.0}$}} & \colorbox[HTML]{67FD9A}{\underline{71.02$_{\pm0.1}$}} &
74.55$_{\pm0.3}$ & 69.20$_{\pm0.1}$ & 67.84$_{\pm0.1}$ \\
Up-Scaling &  & 3.56/10.51 &
\colorbox[HTML]{67FD9A}{85.67$_{\pm0.6}$} & \colorbox[HTML]{67FD9A}{\underline{76.33$_{\pm0.6}$}} & 32.67$_{\pm0.6}$ &
\colorbox[HTML]{67FD9A}{\textbf{44.68$_{\pm0.3}$}} & \colorbox[HTML]{67FD9A}{\textbf{84.16$_{\pm0.1}$}} & 51.15$_{\pm0.4}$ &
\colorbox[HTML]{67FD9A}{\textbf{82.26$_{\pm0.9}$}} & 69.76$_{\pm0.0}$ & 66.40$_{\pm0.1}$ \\
Upcycling & \multirow{-4}{*}{Mix}  & 3.96/10.91 &
\colorbox[HTML]{67FD9A}{\underline{87.00$_{\pm0.0}$}} & \colorbox[HTML]{67FD9A}{69.00$_{\pm0.0}$} & \colorbox[HTML]{67FD9A}{\textbf{51.00$_{\pm1.0}$}} &
\colorbox[HTML]{67FD9A}{\underline{42.58$_{\pm0.3}$}} & \colorbox[HTML]{67FD9A}{81.32$_{\pm0.0}$} & 68.28$_{\pm0.1}$ &
\colorbox[HTML]{67FD9A}{\underline{81.12$_{\pm0.3}$}} & \textbf{70.56$_{\pm0.1}$} & \colorbox[HTML]{67FD9A}{\textbf{68.77$_{\pm0.1}$}} \\

\bottomrule
\end{tabular}
\end{adjustbox}
\end{center}
\caption{Accuracy of the fine-tuned models using trajectories collected from \textit{Qwen3-30B-Instruct}.}
\label{tab:qwen_results}
\end{table*}

\clearpage

\section{Out-of-Domain Generation Example}
\label{app:ood-example}

Examples 1–3 show the generated outputs for a question from FOLIO. The vanilla \textit{Llama-3.1-8B-Instruct} and the deductive Full FT model fail to answer the question correctly, whereas deductive up-scaling produces the correct answer.

\begin{tcolorbox}[float=!htb, colback=red!5, colframe=red!60!black, title=Example 1: Vanilla Llama]
\scriptsize
 Based on the given premises, is the conclusion correct? Please respond with True, False, or Uncertain enclosing in <answer><answer>.\\
Premises: People in this club who perform in school talent shows often attend and are very engaged with school events.\\
People in this club either perform in school talent shows often or are inactive and disinterested community members.\\
People in this club who chaperone high school dances are not students who attend the school.\\
All people in this club who are inactive and disinterested members of their community chaperone high school dances.\\
All young children and teenagers in this club who wish to further their academic careers and educational opportunities are students who attend the school. \\
Bonnie is in this club and she either both attends and is very engaged with school events and is a student who attends the school or is not someone who both attends and is very engaged with school events and is not a student who attends the school.\\
Conclusion: Bonnie performs in school talent shows often.\\

\textbf{Generated Output:}\\
in this case, we can break down the information as follows:\\

1. If Bonnie is a student who attends the school, then she must perform in school talent shows often (from the first premise).\\
2. If Bonnie is not a student who attends the school, then she must be an inactive and disinterested community member (from the second premise).\\
3. If Bonnie is an inactive and disinterested community member, then she must chaperone high school dances (from the fourth premise).\\
4. However, we know that people who chaperone high school dances are not students who attend the school (from the third premise).\\
5. This means that if Bonnie is not a student who attends the school, then she must chaperone high school dances, which contradicts the fact that she is not a student who attends the school.\\
6. Therefore, Bonnie must be a student who attends the school.\\
7. Since Bonnie is a student who attends the school, she must perform in school talent shows often (from the first premise).\\

So, the conclusion is correct: <answer>True</answer>.
\end{tcolorbox}

\begin{tcolorbox}[float=htb, colback=red!5, colframe=red!60!black, title=Example 2: Deductive Full FT Llama]
\scriptsize
\textbf{Generated Output:}\\
 in order to determine if the conclusion is correct, we need to analyze the premises and apply them to Bonnie's situation.\\

From the premises, we know: \\
1. People in this club who perform in school talent shows often attend and are very engaged with school events.\\
2. People in this club either perform in school talent shows often or are inactive and disinterested community members.\\
3. People in this club who chaperone high school dances are not students who attend the school.\\
4. All people in this club who are inactive and disinterested members of their community chaperone high school dances.\\
5. All young children and teenagers in this club who wish to further their academic careers and educational opportunities are students who attend the school.\\

Given that Bonnie is in this club, we can start by looking at the last premise: \\
All young children and teenagers in this club who wish to further their academic careers and educational opportunities are students who attend the school. \\
This implies that if Bonnie is a young child or teenager and wishes to further her academic career and educational opportunities, then she must be a student who attends the school.\\

However, we also know from premise 3 that people in this club who chaperone high school dances are not students who attend the school. \\
If Bonnie is a student who attends the school, then she cannot be a chaperone for high school dances.\\

Now, looking at premise 2: \\
People in this club either perform in school talent shows often or are inactive and disinterested community members.\\
If Bonnie is a student who attends the school, then from premise 1, she must perform in school talent shows often because she is very engaged with school events.\\

However, premise 4 states that all people in this club who are inactive and disinterested members of their community chaperone high school dances. \\
This means that if Bonnie is not a student who attends the school, then she must be an inactive and disinterested member of her community, which would make her a chaperone for high school dances, which contradicts our earlier deduction that if she is a student, she cannot be a chaperone.\\

Considering all premises, we can conclude that if Bonnie is a student who attends the school, then she must perform in school talent shows often. \\
But we also know from premise 3 that if she is a chaperone, she cannot be a student. \\
So, if we consider the possibility that Bonnie is not a student, then she would be an inactive and disinterested member of her community, which would make her a chaperone, but this contradicts our initial assumption that she is not a student.\\

Given the premises, the most consistent conclusion we can draw about Bonnie is that if she is a student who attends the school, then she must perform in school talent shows often. \\
But since we cannot definitively conclude that Bonnie is a student or not based on the given information, we cannot definitively say if she performs in school talent shows often.\\

However, looking closely at premise 5 and the structure of the premises, we can see that if Bonnie is a student who attends the school and is very engaged with school events, then she must perform in school talent shows often. \\
But we cannot conclude this with absolute certainty without more information about Bonnie's status as a student or a non-student.\\

Therefore, based on the given premises, we cannot definitively conclude that Bonnie performs in school talent shows often without assuming that she is a student who attends the school.\\ 
But since the conclusion is about whether Bonnie performs in school talent shows often, and we have a premise that supports this conclusion if she is a student, we can say that the conclusion is True if we assume Bonnie is a student who attends the school and is very engaged with school events.\\

<answer>True</answer>
\end{tcolorbox}

\begin{tcolorbox}[float=!htb, colback=green!5, colframe=green!60!black, title=Example 3: Deductive Up-scaling Llama]
\scriptsize
\textbf{Generated Output:}\\
in order to determine if the conclusion is correct, we need to analyze the premises:\\

1. People in this club who perform in school talent shows often attend and are very engaged with school events.\\
2. People in this club either perform in school talent shows often or are inactive and disinterested community members.\\
3. People in this club who chaperone high school dances are not students who attend the school.\\
4. All people in this club who are inactive and disinterested members of their community chaperone high school dances.\\
5. All young children and teenagers in this club who wish to further their academic careers and educational opportunities are students who attend the school.\\
6. Bonnie is in this club and she either both attends and is very engaged with school events and is a student who attends the school or is not someone who both attends and is very engaged with school events and is not a student who attends the school.\\

From premise 6, we know that Bonnie is in this club and has two possible scenarios:\\
- She both attends and is very engaged with school events and is a student who attends the school.\\
- She is not someone who both attends and is very engaged with school events and is not a student who attends the school.\\

From premise 1, if Bonnie is someone who both attends and is very engaged with school events, then she must perform in school talent shows often.\\

From premise 3, if Bonnie is a student who attends the school, then she cannot chaperone high school dances.\\

From premise 4, if Bonnie is an inactive and disinterested member of her community, then she must chaperone high school dances.\\

From premise 5, if Bonnie is a young child or teenager in this club who wishes to further her academic careers and educational opportunities, then she must be a student who attends the school\\.

Considering all premises, we cannot directly conclude that Bonnie performs in school talent shows often from the given information. However, we can deduce that if Bonnie is a student who attends the school and is very engaged with school events, then she must perform in school talent shows often.\\

But premise 6 does not directly state that Bonnie is a student or that she is not a student. It only gives two possible scenarios for Bonnie's situation.\\

Therefore, the conclusion that Bonnie performs in school talent shows often is not directly supported by the premises, and we cannot be certain of it. \\

<answer>Uncertain</answer>
\end{tcolorbox}

Examples 4–6 show the generated outputs for another question from FOLIO. The vanilla \textit{Qwen-3-8B} and the abductive upcycling model fail to answer the question correctly, whereas the deductive upcycling model produces the correct answer.

\begin{tcolorbox}[float=htb, colback=red!5, colframe=red!60!black, title=Example 4: Vanilla Qwen]
\scriptsize
 Based on the given premises, is the conclusion correct? Please respond with True, False, or Uncertain enclosing in <answer><answer>.\\
Premises: Barutin Cove is a cove named after the Bulgarian settlement of Barutin.\\
Barutin Cove is on the southwest coast of Snow Island.\\
Snow Island, Greenwich Island, and Deception Island are located in the South Shetland Islands.\\
Antarctica is located on the South Shetland Islands. \\
If place A is located in place B and place B is located in place C, then place A is located in place C.\\
Conclusion: Barutin Cove is not located in Antarctica.\\

\textbf{Generated Output:}\\\\
from the premises, we know that barutin cove is on the southwest coast of snow island. snow island is part of the south shetland islands. the south shetland islands are not part of antarctica, but rather located near antarctica. the premises also state that antarctica is located on the south shetland islands, which seems to be a contradiction. however, the correct understanding is that the south shetland islands are a group of islands located near antarctica, not part of it. thus, barutin cove, being on snow island, is not located in antarctica. the conclusion is correct.\\

<answer>True</answer>

\end{tcolorbox}

\begin{tcolorbox}[float=htb, colback=red!5, colframe=red!60!black, title=Example 5: Abductive Upcycling Qwen]
\scriptsize
\textbf{Generated Output:}\\
The premises state that Barutin Cove is on the southwest coast of Snow Island. Snow Island, along with Greenwich Island and Deception Island, is located in the South Shetland Islands. Furthermore, it is stated that Antarctica is located on the South Shetland Islands. \\

However, there's a potential confusion in the premises. The statement ``Antarctica is located on the South Shetland Islands'' is incorrect. In reality, the South Shetland Islands are located in the Southern Ocean, near Antarctica, but they are not part of Antarctica itself. Antarctica is a continent, while the South Shetland Islands are a group of islands that are geographically close to Antarctica but are not part of it.\\

Given the logical structure provided: If place A is located in place B and place B is located in place C, then place A is located in place C. \\

Applying this to the premises: \\
- Barutin Cove is located in Snow Island (A in B).\\
- Snow Island is located in the South Shetland Islands (B in C).\\
- Therefore, by transitivity, Barutin Cove is located in the South Shetland Islands (A in C).\\

But the conclusion states that Barutin Cove is not located in Antarctica. Since the South Shetland Islands are not part of Antarctica, and Barutin Cove is in the South Shetland Islands, it follows that Barutin Cove is not located in Antarctica. \\

However, the premise that ``Antarctica is located on the South Shetland Islands'' is factually incorrect. This might lead to confusion. But based on the logical structure provided and assuming the premises are to be taken as true, even if they contain factual errors, the conclusion that Barutin Cove is not located in Antarctica would still follow from the transitive property of the given premises. \\

Therefore, based on the premises as given, the conclusion is correct.\\

<answer>True</answer>

\end{tcolorbox}

\begin{tcolorbox}[float=htb, colback=green!5, colframe=green!60!black, title=Example 6: Deductive Upcycling Qwen]
\tiny
\textbf{Generated Output:}\\
Okay, let's try to figure this out step by step. So, the question is whether the conclusion ``Barutin Cove is not located in Antarctica'' is correct based on the given premises. Let me break down the information provided.\\

First, the premises say that Barutin Cove is a cove named after the Bulgarian settlement of Barutin. Then, it's located on the southwest coast of Snow Island. Next, Snow Island, Greenwich Island, and Deception Island are all in the South Shetland Islands. Then there's a statement that Antarctica is located on the South Shetland Islands. Wait, that seems a bit off. Wait, actually, the South Shetland Islands are a group of islands located north of the Antarctic Peninsula, and they are part of the Antarctic region, but Antarctica itself is a continent. So maybe there's a mistake here? Let me check that again.\\

The fourth premise says: ``Antarctica is located on the South Shetland Islands.'' That doesn't sound right. Antarctica is a continent, and the South Shetland Islands are a group of islands that are part of the Antarctic region, but they are not part of Antarctica itself. Wait, actually, the South Shetland Islands are considered part of the Antarctic Treaty System, but geographically, they are separate from the continent of Antarctica. So maybe the premise is incorrect? But I have to take the premises as given, even if they might be factually wrong.\\

So, according to the premises, if place A is located in place B and place B is located in place C, then place A is located in place C. That's a transitive property. So, applying that.\\

Barutin Cove is on Snow Island. Snow Island is in the South Shetland Islands. Then, the premise says Antarctica is located on the South Shetland Islands. Wait, that's confusing. If the South Shetland Islands are in Antarctica, then Barutin Cove would be in Antarctica. But the conclusion says it's not. But according to the premises, if the South Shetland Islands are part of Antarctica, then Barutin Cove, being on Snow Island (which is in South Shetland), would be in Antarctica. Therefore, the conclusion that it's not located in Antarctica would be false. But wait, the premise says ``Antarctica is located on the South Shetland Islands.'' That seems like a reversal. Normally, the South Shetland Islands are located in the Antarctic region, but not part of Antarctica. But if the premise states that Antarctica is located on the South Shetland Islands, then that would mean that the South Shetland Islands are part of Antarctica. So, if that's the case, then Barutin Cove is in Antarctica. Therefore, the conclusion that it's not located in Antarctica would be false. But wait, the premises might be conflicting. Let me re-express:\\

Premise 1: Barutin Cove is on Snow Island.

Premise 2: Snow Island is in South Shetland Islands.

Premise 3: South Shetland Islands, Greenwich Island, Deception Island are in South Shetland Islands. Wait, that's redundant. Wait, the third premise says ``Snow Island, Greenwich Island, and Deception Island are located in the South Shetland Islands.'' So, Snow Island is in South Shetland Islands.\\

Premise 4: Antarctica is located on the South Shetland Islands. So, if the South Shetland Islands are in Antarctica, then Barutin Cove is in Antarctica. But the conclusion says it's not. Therefore, the conclusion would be false. But wait, maybe the premise is that the South Shetland Islands are located in Antarctica? Wait, the wording is ``Antarctica is located on the South Shetland Islands.'' That would mean that Antarctica is a part of the South Shetland Islands, which is not correct in reality, but according to the premises, we have to take it as given. So, if Antarctica is located on the South Shetland Islands, then the South Shetland Islands are part of Antarctica. Therefore, Barutin Cove, being on Snow Island (which is in South Shetland), would be in Antarctica. Therefore, the conclusion that it's not located in Antarctica would be false. Therefore, the answer should be False. But wait, maybe I'm misunderstanding the premise. Let me check again.\\

Wait, the fourth premise says ``Antarctica is located on the South Shetland Islands.'' That's a bit confusing. Normally, the South Shetland Islands are located in the Antarctic region, but they are not part of Antarctica. However, if the premise states that Antarctica is located on the South Shetland Islands, that would imply that the South Shetland Islands are part of Antarctica. So, if that's the case, then Barutin Cove is in Antarctica. Therefore, the conclusion that it's not located in Antarctica would be false. Therefore, the answer should be False. But wait, maybe the premise is that the South Shetland Islands are located in Antarctica? But the wording is ``Antarctica is located on the South Shetland Islands,'' which is the opposite. So, if Antarctica is located on the South Shetland Islands, then the South Shetland Islands are part of Antarctica. Therefore, Barutin Cove is in Antarctica. Hence, the conclusion is false. Therefore, the answer is False.\\
<answer>False</answer>

\end{tcolorbox}

\clearpage

\section{Symbolic Generation Example}
\label{app:id_example}

The following examples shows the generated output of answering a symbolic inductive task. The vanilla \textit{Llama-3.1-8B-Instruct} failed to answer the question, while Inductive and Deductive Full FT can generate the correct answer. The vanilla model (Example 7) fails on a inductive task, while inductive (Example 8) and deductive (Example 9) Full FT succeed. Example 8 adopts a pattern-driven, data-oriented perspective, while Example 9 follows a rule-based, hypothesis-testing process, consistent with their respective paradigms. Interestingly, both induced models exhibit self-reflection by revising earlier reasoning. These observations indicate that inducing different reasoning paradigms steers models toward distinct reasoning perspectives, while shared behaviors across induced models demonstrate transferability among these paradigms (\S \ref{sec:symbolic_OOD}).

\begin{tcolorbox}[float=thb, colback=red!5, colframe=red!60!black, title= Example 7: Untrained Llama 3.1 8B]
\tiny
This is a <Inductive> reasoning task. Given the following sequence,\\
$[$'2', '5', '20', '23', '26', '104', '107', '110', '440', '443', '?'$]$ \\
What is the value at the question mark? Please enclose the answer in <answer><answer>, and put all the intermediate reasoning steps in <think><think>.\\

\textbf{Generated Output:}\\
<think>The given sequence appears to be formed by alternating between two different operations: multiplication and addition. To identify the pattern, let's break down the sequence into two separate sequences, one for the numbers that are obtained by multiplication and one for the numbers that are obtained by addition.</think>\\

 Sequence 1 (Multiplication): 2, 20, 104, 440 \\
 Sequence 2 (Addition): 5, 23, 107, 443 \\

<think>Now, let's analyze Sequence 1. The numbers in this sequence are obtained by multiplying the previous term by 10 and then adding 2. This can be represented as: 2 * 10 + 2 = 20, 20 * 10 + 2 = 202, but 202 is not in the sequence, however 20 * 10 + 4 = 204 is not in the sequence either, but 20 * 10 + 2 = 202 is not in the sequence, however 20 * 10 + 4 = 204 is not in the sequence... \\
\textbf{Keeping Repeatitive until hit max number of tokens}
\end{tcolorbox}

\begin{tcolorbox}[float=!htb, colback=green!5, colframe=green!60!black, title=Example 8: Inductive Full FT Llama]
\scriptsize
This is an <Inductive> reasoning task. Given the following sequence,\\
$[$`2', `5', `20', `23', `26', `104', `107', `110', `440', `443', `?'$]$ \\
What is the value at the question mark? Please enclose the answer in <answer><answer>, and put all the intermediate reasoning steps in <think><think>.\\
\textbf{Generated Output:}\\
<think> 
To find the pattern in the sequence, let's first examine the differences between consecutive numbers:\\
5 - 2 = 3, 
 20 - 5 = 15,
 23 - 20 = 3, 
 26 - 23 = 3, 
 104 - 26 = 78, 
 107 - 104 = 3, 
 110 - 107 = 3,
 440 - 110 = 330, 
 443 - 440 = 3\\

From the differences, we can observe a pattern where the difference alternates between 3 and a larger number. The larger numbers are increasing (15, 78, 330), which suggests a pattern of multiplication or addition that increases each time.\\

Looking closer at the larger differences:\\
- 15 * 5.2 = 78 (approximately)\\
- 78 * 4.23 = 330 (approximately)\\

However, a more precise pattern can be observed:\\
15 = 3 * 5,
78 = 15 * 5 + 3,
330 = 78 * 4 + 6\\

But a simpler pattern can be observed if we consider the sequence of operations:\\
Start with 2,
Add 3 to get 5,
Multiply by 4 to get 20,
Add 3 to get 23,
Add 3 to get 26,
Multiply by 4 to get 104,
Add 3 to get 107,
Add 3 to get 110,
Multiply by 4 to get 440,
Add 3 to get 443\\

Following this pattern, the next operation should be to add 3 again, as the pattern alternates between multiplying by 4 and adding 3.\\

So, the next number in the sequence after 443 would be 443 + 3 = 446.
</think>

<answer>446</answer>
\end{tcolorbox}

\begin{tcolorbox}[float=htb, colback=green!5, colframe=green!60!black, title=Example 9: Deductive Full FT Llama]
\scriptsize

\textbf{Generated Output:}\\
<think>
To solve this, let's analyze the given sequence for any patterns or rules that could help us determine the next value. The sequence is: `2', `5', `20', `23', `26', `104', `107', `110', `440', `443', `?'.\\

Looking at the sequence, we can see that it starts with `2' and `5', which are both prime numbers. Then, `20' and `23' are not prime, but `26' is not a prime either, which makes us think that the sequence might not follow a simple pattern of prime numbers.\\

However, if we look closer, we can see that the sequence seems to be formed by adding a certain number to the previous term to get the next term. Let's try to find this pattern.\\

From `2' to `5', the difference is 3. 
From `5' to `20', the difference is 15. 
From `20' to `23', the difference is 3.
From `23' to `26', the difference is 3.
From `26' to `104', the difference is 78.
From `104' to `107', the difference is 3.
From `107' to `110', the difference is 3.
From `110' to `440', the difference is 330.
From `440' to `443', the difference is 3.\\

We can see that the differences are not consistent except for the differences between `2' and `5', `23' and `26', `104' and `107', `107' and `110', and `440' and `443', which are all 3. However, the other differences are not consistent.\\

But if we look at the sequence again, we can see that the numbers `20', `104', and `440' are all multiples of 4, and `23', `26', `107', and `110' are all 3 more than a multiple of 4. \\

So, the pattern seems to be that the sequence is formed by adding 3 to the previous term when the previous term is 3 more than a multiple of 4, and the term before that is a multiple of 4.\\

Let's apply this pattern to find the next term:\\
- The last term before the question mark is `443', which is 3 more than a multiple of 4 (440).\\
- So, the next term should be `443 + 3 = 446'.\\

Therefore, the value at the question mark is 446.\\
</think>

<answer>446</answer>
\end{tcolorbox}

\clearpage
\end{document}